\documentclass[acmsmall,screen,nonacm]{acmart}

\setcopyright{none}
\authorsaddresses{}

\usepackage{amsmath,amsthm,mathtools}
\usepackage{nicefrac}
\usepackage{array}
\usepackage{multirow}
\usepackage{enumitem}
\usepackage{algorithm}
\usepackage{algorithmic}
\usepackage{wrapfig}
\usepackage{subcaption}
\usepackage{pifont}
\usepackage{cleveref}
\usepackage[most]{tcolorbox}
\usepackage{xspace}

\DeclareMathOperator*{\argmax}{arg\,max}

\newcommand{\parahead}[1]{%
    \vspace{0.4em}
    \noindent\textbf{#1}\ \ignorespaces
}

\newcommand{\bea}{BEA\xspace}
\newcommand{\bear}{BEAR\xspace}

\tcbset{enhanced,breakable,boxsep=1mm,left=1.5mm,right=1.5mm,top=1mm,bottom=1mm}
\newtcolorbox{promptbox}[1][]{%
  colback=blue!3,
  colframe=blue!60,
  colbacktitle=black!8,
  coltitle=black,
  fonttitle=\bfseries,
  arc=2mm,
  attach boxed title to top left={xshift=2mm,yshift*=-2mm},
  boxed title style={colframe=black!30},
  #1
}

\title{Mind the Delay: Latency-aware Query Routing for Heterogeneous LLM pools}

\author{Shivam Patel}
\authornote{Equal contribution.}
\affiliation{%
  \institution{Carnegie Mellon University}
  \city{Pittsburgh}
  \state{Pennsylvania}
  \country{USA}}

\author{Akaash R. Parthasarathy}
\authornotemark[1]
\affiliation{%
  \institution{Carnegie Mellon University}
  \city{Pittsburgh}
  \state{Pennsylvania}
  \country{USA}}

\author{Akash Dhasade}
\affiliation{%
  \institution{Carnegie Mellon University}
  \city{Pittsburgh}
  \state{Pennsylvania}
  \country{USA}}  

\author{Ankur Mallick}
\affiliation{%
  \institution{Microsoft}
  \country{USA}}

\author{Gauri Joshi}
\affiliation{%
  \institution{Carnegie Mellon University}
  \city{Pittsburgh}
  \state{Pennsylvania}
  \country{USA}}

\renewcommand{\shortauthors}{Patel et al.}

\begin{document}

\begin{abstract}
Large language model (LLM) query routers typically assign each query to a suitable model that balances response accuracy and monetary cost. Subsequently, separate system-level load balancers based on policies such as join-the-shortest-queue or round-robin distribute the queries across replicas of each model to reduce latency. However, decoupling accuracy-cost and latency considerations can lead to suboptimal routing decisions. We introduce the \textit{Batch Evolution Aware} (\bea) estimator to provide a fine-grained estimate of the latency, that is, the time-to-first-token (TTFT) experienced by an arriving query. Unlike existing load balancers that rely on coarse signals such as queue lengths, per-model processing rates, or historical average latencies, \bea replays the serving framework's token-batch formation using a snapshot of an instance's current workload, stopping when the incoming query's first token is generated. Taking into account the compute and memory requirements of the prefill and decode phases of language queries as well as the contention between multiple queries that are served in the same batch, \bea predicts TTFT within an error of $\sim\!5\%$. Each prediction takes about $0.1$\, ms, which is fast enough to run on the critical routing path. When workload snapshots are unavailable, we also design an analytical framework for a model instance under a limited processor-sharing scheme that calculates expected TTFT in closed form and captures the qualitative trend of latencies as the load increases.
The TTFT estimates produced by \bea are then used in our latency-aware router, \bear, which simultaneously balances accuracy, cost, and latency to select the best model instance from a heterogeneous pool of LLMs. Against seven baselines, \bear improves the utility of responses delivered within their targets by $63\%$ on average over existing latency-aware routers, and by $30.5\%$ over the best baseline at each arrival rate.
\end{abstract}


\maketitle

\section{Introduction}
\label{sec:introduction}

Modern Large language models (LLMs) differ widely in their response quality and inference costs for generating responses to inference queries, even within a single model family. Query routers learn the mapping between query requirements and model capabilities to better utilize the heterogeneity in the model pool. For each incoming query, the router predicts the accuracy of each available model in answering that query and the incurred inference cost, consequently routing queries based on a chosen accuracy--cost trade-off. Easy queries are often routed to small and inexpensive models, whereas difficult queries are routed to large and capable models. A substantial body of work learns these predictions from historical query-model evaluations \cite{chen2024frugalgpt,ding2024hybridllm,ong2025routellm,hu2024routerbench,jitkrittum2026universal,patel2025proxrouterproximityweightedllmquery,ding2025bestroute,dekoninck2025a_unified_routing_cascading,askin2026federaterouterlearninglanguage}.

In practical deployments, each language model in the pool is served by finite and shared hardware resources. 
Models are hosted on GPUs and queries are served using frameworks such as vLLM \cite{10.1145/3600006.3613165_vllm} or Sarathi-Serve \cite{10.5555/3691938.3691945_sarathi}. We refer to this model-framework combination as a \emph{model instance}. Serving frameworks process multiple queries concurrently. Thus, the delay experienced by a query depends on the real-time workload at the model instance due to resource sharing. A popular measure of inference latency for interactive applications is the \emph{time-to-first-token} (TTFT), i.e., the time from a query's arrival until the first response token is produced. Serving systems are commonly designed and evaluated against TTFT targets \cite{distserve,10.5555/3691938.3691945_sarathi} across a variety of query distributions.

We study delay-constrained routing where each query carries a TTFT target, and the router selects, among the model instances expected to meet that target, the one with the highest predicted accuracy--cost utility. 
Our formulation critically uses the latency estimator to predict a feasible set of latency-compliant model instances for each query. If the latency predictor overestimates TTFT, the router discards capable instances that would have satisfied the chosen latency targets. If it under-estimates TTFT, the router would dispatch queries to instances which would violate the desired latencies. Thus, the core question we seek to address is: 

\begin{center} \emph{When a query arrives, how can a router accurately predict the TTFT at each model instance and distinguish between feasible and infeasible instances, while remaining lightweight enough to operate on the critical path of every real-time routing decision?}
\end{center}

\subsection{Limitations of existing approaches}
\label{subsec:intro-prior-approaches}

Two bodies of work separately address accuracy--cost based routing, and latency minimization across deployed hardware. Only a few recent routing paradigms attempt to jointly address latency and accuracy--cost optimization. Below, we summarize the literature in these broad directions and defer a more complete discussion to \Cref{subsec:related-work}.

\parahead{Accuracy--cost based routers.}
The first line of work is in machine learning literature, where query routers aim to balance response accuracy against inference costs \cite{ong2025routellm,ding2024hybridllm,jitkrittum2026universal,patel2025proxrouterproximityweightedllmquery,ding2025bestroute,dekoninck2025a_unified_routing_cascading}. These routers are trained on historical (query, model) pair evaluation data to produce accuracy and cost estimates, and these estimates are used to make routing decisions at inference time. However, since both accuracy and inference cost are attributes of the language models themselves rather than hardware on which they are deployed, these approaches do not consider workload congestion and the scheduling policy used by the serving framework, and thus they are latency-agnostic. 

\begin{figure}[t]
    \centering
    \includegraphics[width=0.9\linewidth]{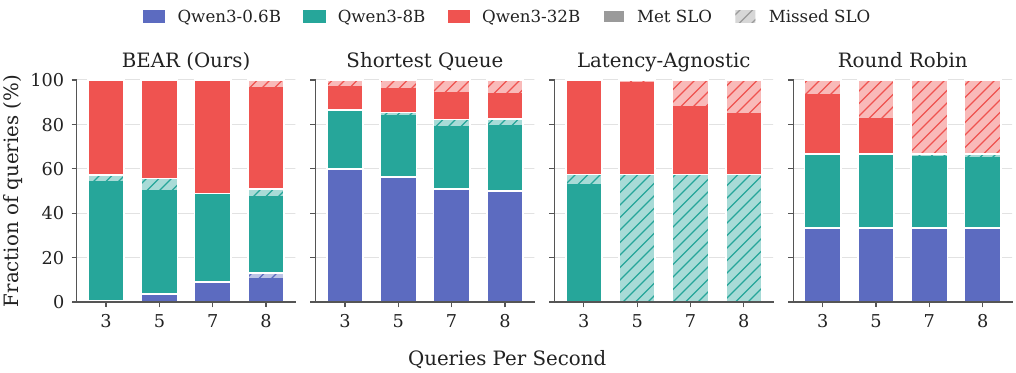}
    \caption{\small \textbf{Routing composition and TTFT target attainment across arrival rates.}
    Solid and hatched portions denote queries that meet and miss their TTFT targets, i.e., their service level objectives (SLOs). \bear adapts model assignments as load grows, meeting most targets while still using the capable model instances. Shortest Queue also meets most targets but relies heavily on smaller model instances, whereas Latency-Agnostic and Round Robin keep nearly constant allocations and miss targets more often.}
    \label{fig:routing_composition}
\end{figure}

\parahead{Load balancing and job scheduling in serving systems.} 
The second line of work is in systems-focused research to reduce the latency experienced by queries for a fixed given language model and hardware configuration. Dispatching policies such as Round Robin and Join-the-Shortest-Queue
\cite{harchol-balter2013performance}, as well as LLM-specific load balancing policies \cite{10.1145/3721146.3721947_performance_aware_load_balancing,10.5555/3691938.3691948_lluminix,da2025blockbalancingloadllm,zhang2025lmdeploy}, distribute queries across multiple instances of the same model to obtain varying latency objectives and desired serving behavior for downstream applications. Alternatively, scheduling policies reorder queries within a chosen model instance based on latency constraints, predicted output lengths and remaining decode workload to minimize deadline violations and average latencies \cite{huang2025sloawareschedulinglargelanguage,aiops2024qiu_efficient_llm_scheduling_with_proxy_model,fu2024efficient_scheduling_learning_to_rank,shahout2025dontstopmenow,10.1145/3698038.3698523_qlm}. All the above methods focus on varying serving capacities of instances of the same model and do not consider the accuracy and cost characteristics of different models. In a heterogeneous model pool, it is essential to consider the predicted accuracy and cost of a query assigned to a model and jointly optimizing accuracy, cost, and latency. In our experiments (\Cref{fig:routing_composition}), we observe that join-the-shortest-queue, which is latency-aware but agnostic to accuracy and cost, satisfies query TTFT targets at a high rate, but it does so by relying heavily on the smallest model and underutilizing more capable model instances.

\parahead{Joint accuracy--cost--latency optimizing routers.}
SCORE \cite{lakha2025faster_cheaper_justasgood} and RouteBalance \cite{da2026routebalancefusedmodelrouting} consider a latency-aware routing objective and the throughput of each model instance to estimate the average latency experienced by queries. SCORE estimates latency as the queueing delay plus the predicted output length (in tokens) multiplied by the per-token processing time of that model. RouteBalance combines calibrated time-per-output-token predictions with estimated output lengths. Both of these approaches use coarse throughput-based estimates of generation latencies. However, in practice, queries contend for compute and memory resources and are served in batches rather than sequentially. Thus, finer-grained latency estimation is needed to improve model resource utilization and provide better routing performance.

\subsection{Challenges in Estimating Time-to-first-token (TTFT) Latency}
\label{subsec:intro-challenges}

We identify key challenges in estimating TTFT for queries, and highlight their consequences. 

\parahead{Real-time serving state affects TTFT.}
Unlike model accuracy and cost, which can be learned from past evaluations, token counts, and per-token inference costs, estimating TTFT depends heavily on the current state of workload at the model instance when a new query arrives. The queries currently at model instances that are in the decode stage (explained below), their remaining response lengths, and number of queued queries in the prefill stage (explained below) affect the latencies experienced by a new query arrival. Hence, the same query arriving at the same model instance can experience very different TTFTs based on the state and workload congestion at the model instance. 

\parahead{Cross-query resource contention in token batches.}
Autoregressive generation of responses to queries using LLMs proceeds in two distinct phases (\Cref{subsec:background-inference}). In the \emph{prefill} phase, the model processes all prompt (query) tokens in parallel, stores their attention keys and values in GPU memory (the \emph{KV cache}), and produces the first output token. In the \emph{decode} phase, the response (output) tokens to the query are generated one at a time. Each new token uses \emph{attention} mechanisms to combine information from the cached keys and values of all preceding tokens, resulting in growing computational intensity of a decode step as the query's preceding context length grows during response generation.

Modern serving frameworks interleave these phases across queries using continuous batching \cite{280922_orca,10.1145/3600006.3613165_vllm}, processing one \emph{token batch} per iteration. With chunked prefills \cite{10.5555/3691938.3691945_sarathi}, prompt tokens of newly admitted queries and output tokens of decoding queries are processed in the same batch without requiring separate batches for prefill and decode computations. A new query's prefill stage therefore competes with the decode workload of active queries at the model instance, resulting in slower generation in the presence of high number of actively decoded queries or long contexts (\Cref{subsec:background-serving}). The queue length and the number of remaining prefill tokens do not reflect this decode interference, and two model instances with identical queued queries and prefill token counts can lead to significantly different TTFT latencies. For example, a naive estimator that divides the pending prompt tokens in an instance by the rate at which the instance processes prompt tokens (its prefill throughput) has an $85\%$ mean absolute percentage error (MAPE) on TTFT for Qwen3-0.6B on one H100 GPU (\Cref{fig:ttft_estimator_comparison}).

\parahead{Predicting future workload of existing queries.}
The token batch composition throughout to the new query's first output token utilizes the number of sequences that are being actively decoded at the model instance. This depends on the remaining output lengths of the active sequences, which are not accurately known when estimating TTFT latencies for new queries. Furthermore, the remaining output lengths of the sequences that are being actively decoded and the output lengths of queued sequences affect the KV-cache memory consumption, which may affect the waiting times for a new query before its prefill computation can begin.

\parahead{Latency estimation lies on the critical path.}
The router must produce an estimate for each arriving query at every candidate model instance before making a routing decision. Since routing cannot be performed until the accuracy, cost, and latency predictions for a given query are available, the estimation time directly contributes to a query's latency. 
Existing simulators such as Vidur \cite{MLSYS2024_b74a8de4_vidur} model execution at individual kernel operator level in order to accurately evaluate deployment configurations offline. While they are well suited for offline analysis, such approaches become too expensive and cumbersome for performing real-time routing decisions. A practical estimator must strike a balance between accurate predictions and coarse granularity, executing predictions on a fast time scale to ensure benefits of latency-aware routing are not offset by the estimation costs themselves.

\subsection{Our approach}
\label{subsec:intro-approach}

\parahead{Key insight: Lightweight simulation of the evolving inference state.}
Aggregate statistics such as queue length, pending prompt tokens, or average throughput ignore the composition of the token workload at a model instance, which is crucial for determining how the serving framework \cite{10.1145/3600006.3613165_vllm,10.5555/3691938.3691945_sarathi} forms the next several token batches. The inference framework can be uniquely represented as a deterministic set of rules applied to the token counts of resident requests. 
If the remaining number of prompt and output tokens of each request at the instance are known, the sequence of token batches preceding a new query's first output token can be fully determined. At routing time, only the remaining output lengths for queries need to be predicted. 
A router can therefore estimate the TTFT by replaying the scheduler forward from the current state, rather than modeling the workload summary to estimate latencies.

\parahead{\bea latency estimator.}
We build the \emph{Batch Evolution Aware} (\bea) latency estimator based on the aforementioned observation (\Cref{subsec:sfs-latency-estimator}). After every token iteration, each model instance publishes a snapshot of its resident requests (actively decoded and already queued) and the number of prompt tokens yet to be processed for each query. When a new query arrives, the router augments the snapshot with predicted output lengths for the resident requests and the new query. It then replays the serving framework's batching and scheduling policies, including its token budget, admission limits, and KV-cache memory constraints to obtain a sequence of batches and the comprising prefill and decode token workloads.
We then estimate the TTFT as the sum of the predicted processing times of the simulated token batches, up to the batch that completes the new query's prefill and produces its first output token.

We make two design choices which improve the practicality of our latency estimator. 
First, we model processing time per token batch rather than per GPU kernel operator. 
Our token batch processing time model comprises of four terms, each corresponding to a separate computational aspect of inference: a fixed per-batch overhead, a fixed per-token computation cost for linear and MLP (multi-layer perceptron) layers, a decode-attention cost that grows with the context length of each query, and a prefill-attention cost that grows with both the number of prompt tokens in the batch and the length of the preceding context. We calibrate the coefficients for each of the computational terms from observed batch times for each model instance. Second, we stop the batch evolution replay at the new query's first output token and do not simulate resident requests to completion. Only the batch-construction step of the replay depends on the serving policy, so adapting the estimator to another serving policy only requires changing the scheduling simulator.

\begin{figure}[t]
    \centering    \includegraphics[scale=0.55]{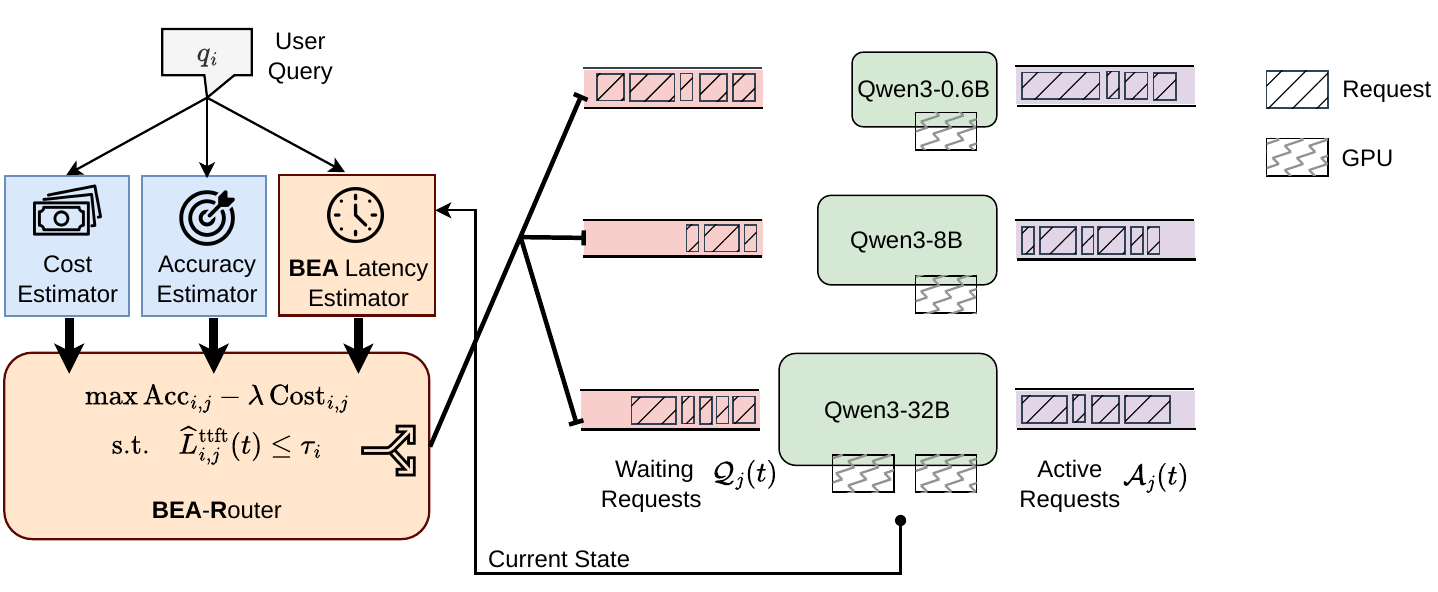}
    \caption{\textbf{Latency-aware routing with \bear.} 
    Upon receiving a user query $q_i$, the expected accuracy, cost and TTFT latency are estimated for each model instance. Our \bea estimator predicts TTFT by replaying the serving framework's token-batch formation using a snapshot of the current workload, i.e., waiting ($\mathcal{Q}_j(t)$) and active ($\mathcal{A}_j(t)$) requests.
    Our router, \bear, then selects the model instance with the highest predicted accuracy--cost utility among the feasible set of model instances that satisfy the query's TTFT constraints.}
    \label{fig:high-level-diagram}
\end{figure}

\parahead{\bear Router.}
The \emph{Batch Evolution Aware Router} (\bear) combines \bea latency estimates with accuracy and cost predictions to perform routing~(\Cref{fig:high-level-diagram}). For each query, it selects the instance with the highest predicted accuracy--cost utility among the feasible set of model instances that satisfy a query's TTFT constraints. We also study and evaluate a Lagrangian relaxation of the constrained routing objective, which penalizes the TTFT latency and permits a controllable trade-off between accuracy--cost utility and latency (\Cref{sec:formulation}).

\parahead{Analytical model for Autoregressive Generation.}
The \bea estimator requires real-time workload snapshots, which may not always be available to the router under certain settings. For this scenario, we analytically model the autoregressive generation as a limited processor sharing (LPS) \cite{zhang2009limited_processor_sharing} scheme, in which up to $k$ queries are served concurrently and the service capacity of the instance is jointly shared by each active query, and further queries wait in a first-come-first-served (FCFS) queue (\Cref{subsec:lps_queueing_delay}). The concurrency limit reflects the KV-cache memory bound on continuous batching, and capacity sharing captures the resource sharing and contention among actively generated responses. Under Poisson arrivals and exponential service time assumptions, the number of queries at the instance evolves as a birth--death chain process of an $M/M/1$ queue, which yields the expected queueing delay in closed form as $\rho^{k}/(\mu-\alpha)$, where $\alpha$ is the arrival rate, $\mu$ the service rate of the model instance, and $\rho=\alpha/\mu$ is the utilization. For large $k$, the factor $\rho^{k}$ reduces the average queueing delay (and consequently TTFT as well), and increasing utilization ($\alpha\!\to\!\mu$) results in growing queueing delays. For Qwen3-0.6B language model hosted on one H100 GPU, the measured TTFT follows the analytical prediction with $k=25$, close to the measured average of $25.7$ active sequences (\Cref{fig:theoretical_vs_empirical_ttft_queueing_lps}).

\subsection{Main results}
\label{subsec:intro-results}

We evaluated \bear on a pool of three vLLM-served model instances: Qwen3-0.6B and Qwen3-8B running on one H100 GPU each, and Qwen3-32B hosted on two H100 GPUs with tensor parallelism. 
Queries are drawn from four tasks \cite{alpaca_dataset,yang2018hotpotqa,huang-etal-2021-efficient_govreport_summarization,fan-etal-2018-hierarchical_writing_prompts_dataset} that together cover short and long prompts paired with short and long responses, and follow the  Poisson and two-state Markov-modulated Poisson arrival processes \cite{fischer1993mmpp}. 
We compare against seven baselines, including two joint quality--cost--latency routers (RouteBalance and SCORE), a latency-agnostic quality--cost router, and four load-balancing policies (round robin, join-the-shortest-queue, LMDeploy~\cite{zhang2025lmdeploy}, and vLLM's Semantic Router~\cite{liu2026vllmsemanticroutersignal}). 
Our primary evaluation metric is \emph{OnTimeUtility}, the average accuracy--cost utility per query with zero utility assigned to queries that miss their TTFT target (\cref{eq:ontime_utility}).

\parahead{Latency prediction.}
Our evaluations show that the \bea estimator predicts TTFT with $\approx\!5\%$ MAPE while its batch processing time model has $\approx\!4\%$ MAPE across token-batch compositions on the Qwen3-0.6B model instance~(\Cref{fig:ttft_estimator_comparison,fig:batch_processing_time}).
Furthermore, the estimation overhead of $\approx\!10^{-4}$\,s (\Cref{fig:simulation_latency}) is negligible relative to the TTFT targets greater than $150$\,ms used in our experiments.

\parahead{Routing performance.}
\bear attains the highest OnTimeUtility at each arrival rate we evaluate (\Cref{fig:ontimeutility_vs_qps}). It improves OnTimeUtility by $63\%$ on average over RouteBalance and SCORE, with a maximum improvement of $74\%$ at $8.9$ queries per second. Additionally, \bear exhibits an average improvement rate of $30.5\%$ compared to the best baseline at each arrival rate.

\parahead{Interpreting performance improvements.}
As the load increases, \bear routes a growing fraction of queries to the smaller and faster Qwen3-0.6B model instance, but continues to use the larger models whenever their predicted TTFT meets the target (\Cref{fig:routing_composition}). The latency-agnostic router and round robin keep nearly fixed allocations and violate latency targets more often as load grows, while join-the-shortest-queue meets targets but under-utilizes more capable model instances. 
The \bea estimator contributes directly to these gains.
Replacing the \bea estimator with the prefill-throughput estimator while keeping the rest fixed gives comparable OnTimeUtility at low arrival rates. However, it suffers from an increasingly lower OnTimeUtility as the load increases~(\Cref{fig:prefill_throughput_ablation}).

\parahead{Utility--latency trade-off and robustness.}
In the Lagrangian relaxation (\cref{eq:lagrangian_routing_objective}) of the latency constrained routing objective (\cref{eq:constrained_routing_objective}), varying the latency penalty traces a utility--latency trade-off curve, allowing a choice of desired operating point~(\Cref{fig:utility_vs_latency}). 
Across this trade-off, \bear improves performance over the examined baselines.
\bear also retains the highest OnTimeUtility under bursty arrivals~(\Cref{fig:query_arrival_distribution_ablation}). 
Finally, when considering stale snapshots, its OnTimeUtility stays close to the zero-delay value when workload snapshots are up to $400$\,ms stale, and decreases by only $7\%$ when they are up to $1$\,sec stale~(\Cref{fig:snapshot_staleness}).

\subsection{Key Contributions}
\label{subsec:intro-contributions}

\begin{itemize}[leftmargin=*]
    \item \textbf{A workload- and scheduler-aware TTFT estimator.} We design the \bea estimator which accurately models the prefill vs. decode interference in token batches. 
    The \bea estimator operates by replaying the serving framework's batching and scheduling policies from a snapshot of the resident requests on the instance.
    Uniquely, it estimates the processing time of each simulated batch using four calibrated coefficients, enabling it to predict TTFT with about $5\%$ MAPE. The prediction overhead remains small, taking only $10^{-4}$\,s per estimate.
    \item \textbf{An analytical model of Autoregressive Generation.} For settings in which the instance's state cannot be observed, we model a model instance as a limited processor sharing queue. Our model captures the memory-bounded concurrency and capacity sharing of continuous batching, giving us the expected queuing delay of an arriving query in closed form, from the instance's arrival rate, service rate, and concurrency. Empirically, the measured TTFT matches the model's prediction at a concurrency level consistent with the measured number of active sequences.
    \item \textbf{A latency-aware router and its evaluation.} We formulate routing across heterogeneous model instances under per-query TTFT targets and instantiate our \bea estimator for routing as \bear. Against seven baselines, \bear improves OnTimeUtility by $63\%$ on average over joint accuracy--cost--latency routers and by $30.5\%$ over the best baseline at each arrival rate. 
    Through extensive ablations, we show that the gains of \bear persist across bursty arrivals and stale workload snapshots. 
\end{itemize}

\parahead{Paper organization.}
The rest of the paper is organized as follows. \Cref{sec:serving-background} reviews autoregressive inference and LLM serving frameworks and discusses related work. \Cref{sec:formulation} formalizes the routing problem and the evaluation metrics. \Cref{sec:latency-estimation} presents the \bea estimator and the LPS model. \Cref{sec:experiments} evaluates \bear, and \Cref{sec:conclusion} concludes our work.

 \section{Preliminaries and Related Work}
\label{sec:serving-background}
We first review how transformer models generate text autoregressively (\Cref{subsec:background-inference}), and then how serving frameworks batch, allocate memory for, and schedule this computation across many queries (\Cref{subsec:background-serving}). We finally present related work in \Cref{subsec:related-work}.

\subsection{Autoregressive inference with transformers}
\label{subsec:background-inference}

\parahead{Autoregressive generation.}
Most transformer-based language models generate text autoregressively, predicting one token in each iteration by attending to the input prompt and all previously generated tokens \cite{NIPS2017_3f5ee243_attention_is_all_you_need}.
Consider a query $q$ with $\textrm{tok}^{\textrm{pre}}_{q}$ denoting the number of prompt tokens and $\textrm{tok}^{\textrm{dec}}_{q}$ denoting the number of output tokens to be generated.
Let $x_1, x_2, \dots$ denote the tokens of the sequence associated with $q$, so that $x_1, \dots, x_{\textrm{tok}^{\textrm{pre}}_{q}}$ is the prompt. A language model $p_\theta$ generates the response one token per iteration, each conditioned on all preceding tokens:
\begin{equation}
    x_{s+1} \sim p_\theta\left(\cdot \mid x_1, \dots, x_s\right), \qquad s = \textrm{tok}^{\textrm{pre}}_{q}, \dots, \textrm{tok}^{\textrm{pre}}_{q}+\textrm{tok}^{\textrm{dec}}_{q}-1.
    \label{eq:next_token}
\end{equation}
The generated token is appended to the sequence, and generation stops when an end-of-sequence token is produced or the maximum length is reached. Since each token depends on its predecessors, output tokens must be generated sequentially, with one forward pass through the model per token.
Although other paradigms exist, such as diffusion-based generation and non-autoregressive refinement \cite{gu2018nonautoregressive,lee-etal-2018-deterministic_non_autoregressive,li2022diffusionlm}, we focus on the widely-used attention-based autoregressive setting.

\parahead{Attention.}
A transformer consists of a stack of $n_{\textrm{layer}}$ layers, each combining a self-attention block with a feed-forward (MLP) block. After the final layer, a single output projection (the ``LM head'') maps the representation of the last position to the next-token distribution.
Attention is the only operation that combines information across positions; all other operations act on each token independently.
Let $\mathbf{u}_s \in \mathbb{R}^{d}$ denote the input to a layer at token position $s$, which corresponds to the embedding of token $x_s$ for the first layer, and the previous layer's output otherwise. Attention projects $\mathbf{u}_s$ into a query $\mathbf{q}_s = \mathbf{W}_{Q}\mathbf{u}_s$, a key $\mathbf{k}_s = \mathbf{W}_{K}\mathbf{u}_s$, and a value $\mathbf{v}_s = \mathbf{W}_{V}\mathbf{u}_s$, each in $\mathbb{R}^{d_{\textrm{h}}}$, using learned weight matrices. (The bold attention query $\mathbf{q}_s$ should not be confused with a user query $q$.) Stacking the keys and values of all positions up to $s$ row-wise into $\mathbf{K}_s = [\mathbf{k}_1, \dots, \mathbf{k}_s]^{\top} \in \mathbb{R}^{s \times d_{\textrm{h}}}$ and $\mathbf{V}_s = [\mathbf{v}_1, \dots, \mathbf{v}_s]^{\top} \in \mathbb{R}^{s \times d_{\textrm{h}}}$, the output at position $s$ is a weighted average of these values:
\begin{equation}
    \mathbf{o}_s = \mathbf{V}_s^{\top}\, \textrm{softmax}\!\left(\frac{\mathbf{K}_s\, \mathbf{q}_s}{\sqrt{d_{\textrm{h}}}}\right),
    \label{eq:attention}
\end{equation}
where the softmax normalizes the $s$ query--key scores into weights that sum to one.
Because $\mathbf{K}_s$ and $\mathbf{V}_s$ contain only positions $(1, \dots, s)$, each position attends only to itself and earlier tokens (called \emph{causal masking}).
In practice, each layer uses \emph{multi-head} attention: several such operations (heads) run in parallel with separate weights, and their outputs are concatenated and passed through the MLP block to produce the next layer's input $\mathbf{u}_s$.

\parahead{KV cache.}
Under causal masking, the key and value at each position depend only on the tokens up to that position, so they do not change once computed.
Serving frameworks therefore store them in a \emph{KV cache} instead of recomputing them in every iteration.
For each sequence, the KV cache holds the matrices $(\mathbf{K}_s, \mathbf{V}_s)$ in every layer and for every key/value head, and each newly processed token appends one row to each. Nothing else is cached; in particular, the query $\mathbf{q}_s$ is only needed to compute the output at its own position $s$.

\parahead{Prefill and decode phases.}
Generating a response to query $q$ proceeds in two computational phases~\cite{MLSYS2023_c4be71ab_efficiently_scaling_transformer_inference}. In the \emph{prefill} phase, all $\textrm{tok}^{\textrm{pre}}_{q}$ prompt tokens are processed in parallel, evaluating \cref{eq:attention} for all prompt positions at once, to compute their keys and values and populate the KV cache.
The phase concludes by applying the LM head at the last prompt position to generate the first output token.
In the \emph{decode} phase, the remaining $\textrm{tok}^{\textrm{dec}}_{q}-1$ output tokens are generated one at a time: each step computes the query, key, and value of only the newest token, appends its key and value to the cache, and evaluates \cref{eq:attention} against all cached keys and values.
Since prompt tokens are processed in parallel, the prefill phase is typically compute-bound, whereas the decode phase is typically memory-bound, as each decode step must access cached KV states of all preceding tokens~\cite{10.5555/3691938.3691945_sarathi,10.1145/3600006.3613165_vllm}.

\subsection{LLM serving frameworks}
\label{subsec:background-serving}

\parahead{Batch composition and generation throughput.}
Batching of computational workload for generating responses to queries is a crucial aspect of utilizing the parallel computation capacity of GPUs. Naive sequence-level batching collects queries and generates new tokens for each query in the batch until completion. This leads to inefficient hardware utilization as shorter sequences in the batch need to wait for longer sequences to finish generation. To overcome this, modern serving frameworks utilize continuous batching (ORCA \cite{280922_orca}, vLLM \cite{10.1145/3600006.3613165_vllm}, etc.), where new prompts are admitted in the generation stage at token iteration (often simply denoted as ``token batch'') boundaries, unlike sequence-level batch boundaries in naive batching. This implies that autoregressive generation for a new sequence can begin while other sequences are still being generated.
Continuous batching leads to better hardware utilization and higher throughput.

\parahead{Memory allocation.}
The KV cache of each sequence grows with every generated token, and the GPU memory available for the KV cache limits how many sequences can be served concurrently. Efficient KV cache memory allocation is therefore crucial for achieving high throughput.
Early solutions, such as ORCA \cite{280922_orca}, reserve memory for the KV cache up to the maximum possible context length of each sequence being served. 
This leads to over-provisioning since most sequences do not achieve the maximum context length, resulting in suboptimal hardware utilization and reduced throughput. PagedAttention (used in vLLM) \cite{10.1145/3600006.3613165_vllm} applies virtual memory-style paging to the KV cache, avoiding large contiguous pre-allocation and instead allocating fixed-size KV cache blocks to sequences dynamically as generation proceeds.

\parahead{Scheduling mechanisms.}
The heterogeneous compute and memory requirements of the prefill and decode phases give rise to different policies for composing token batches.
\begin{itemize}[leftmargin=*]
\item \emph{Prefill-prioritizing} schedulers, such as the scheduler in earlier versions of vLLM \cite{10.1145/3600006.3613165_vllm}, process the prompts of newly arrived sequences as soon as memory permits, computing all $\textrm{tok}^{\textrm{pre}}_{q}$ prompt tokens of a new query $q$ in a single token batch. This reduces the TTFT of new queries but stalls the decode phase of ongoing sequences, since no ongoing sequence generates its next token until the entire prefill of $q$ completes.
\item \emph{Decode-prioritizing} schedulers instead advance ongoing decodes before admitting new prompts, which keeps inter-token latency low but increases the waiting time, and hence the TTFT, of new queries.
\item \emph{Chunked prefills}~\cite{10.5555/3691938.3691945_sarathi} balance these two extremes by splitting the $\textrm{tok}^{\textrm{pre}}_{q}$ prompt tokens of a query $q$ into smaller chunks that are processed over multiple token batches, where each token batch is limited to a budget of $\kappa$ tokens.
These chunks can be scheduled in prefill-only batches that interleave with decode batches, or co-scheduled with ongoing decodes within the same token batch.
In the latter case, each token batch contains one decode token for each of the $n^{\textrm{dec}}$ active decode sequences, and the remaining budget of $\kappa - n^{\textrm{dec}}$ tokens goes to prefill chunks.\footnote{Following FCFS order, we assume the remaining budget goes entirely to the prefill of the earliest-arrived incomplete sequence, so later arrivals do not affect the prefill of earlier requests.} 
\end{itemize}
Alternatively, prefill and decode can be disaggregated onto separate GPUs \cite{distserve}, which removes interference between the two phases at the cost of transferring the KV cache between them. 
In this work, we focus on the co-located setting, where prefill and decode share the same GPU, as in the default configurations of widely used serving frameworks.
Our latency estimator simulates the token-batch composition induced by the serving framework's scheduling policy, capturing these interactions to produce accurate TTFT estimates for queries.

\subsection{Related work}
\label{subsec:related-work}
Multiple research directions tackle efficient response generation through varying approaches. Accuracy and cost based language query routing performs model selection to maximize the collective performance on queries. 
Alternatively, systems-focused research emphasizes latency-constrained response generation through improved scheduling, workload-aware load balancing and queue management policies. Crucially, a separate examination of these directions does not provide a structured understanding of the joint accuracy--cost--latency tradeoffs arising in response generation. Our work connects these directions by incorporating workload-dependent TTFT latency estimates into routing decisions across heterogeneous model instances. 
Below, we expand on concurrent research directions focusing on efficient response generation with multiple model instances.

\parahead{LLM routing for quality--cost trade-offs.}
LLM routers leverage differences in model capabilities and inference costs to choose appropriate model for each test query. FrugalGPT~\cite{chen2024frugalgpt} builds cost-effective model cascades for sequential inference, while HybridLLM~\cite{ding2024hybridllm} routes queries between a small and a large model based on the predicted query difficulty and the desired response quality, whereas RouteLLM~\cite{ong2025routellm} trains query routers using human preference comparisons between model responses. RouterBench~\cite{hu2024routerbench} provides a framework for evaluating quality--cost routing, including nearest-neighbor approaches, while Universal Model Routing~\cite{jitkrittum2026universal} presents and theoretically examines cluster-based routing strategies that can adapt to a time-varying pool of language models. ProxRouter~\cite{patel2025proxrouterproximityweightedllmquery} utilizes proximity-weighted performance estimation in query embedding space to improve the robustness of nonparametric routers to out-of-distribution test queries. Additional approaches employ item response theory to model query difficulty and model abilities~\cite{song-etal-2025-irt_router,chen2025learningcompactrepresentationsllm_irt_net}, or utilize compact model embeddings for predicting model performance and selection \cite{zhuang2025embedllm,patel2026locuslowdimensionalmodelembeddings}. 
These approaches primarily improve quality--cost based model selection for queries without explicitly modeling time and workload dependent latencies at candidate model instances.

\parahead{Dynamic scheduling and queue management.}
Inference serving for queries with heterogeneous latency requirements and varying priorities motivates scheduling policies which account for generation constraints, limited resource availability and cross-query workload interference during autoregressive generation. 
SLO-aware scheduling~\cite{huang2025sloawareschedulinglargelanguage} assigns request priorities using latency targets and the query input and estimated output lengths. QLM~\cite{10.1145/3698038.3698523_qlm} combines waiting-time estimation with queue management to orchestrate request-group scheduling, load balancing, eviction, and model swapping for improved service level objective (SLO) attainment. Llumnix~\cite{10.5555/3691938.3691948_lluminix} dynamically reschedules running requests through live migration of their execution states, improving load balancing, isolation, and resource utilization across model instances. Queueing-theoretic studies characterize throughput-optimal scheduling, resource allocation, and the challenges introduced by batching, uncertain service times, and changing memory requirements~\cite{li2025throughputoptimalschedulingalgorithmsllm_li,10.1145/3771574_optimal_algorithms_hegde,mitzenmacher2025queueingpredictionsllmschallenges}. These directions broadly focus on improved generation and inference execution with fixed model choices for queries.

\parahead{Workload-aware routing and latency modeling.}
Workload-aware dispatching of queries across serving instances complements priority based scheduling policies within each instance. Performance-aware load balancing~\cite{10.1145/3721146.3721947_performance_aware_load_balancing} uses reinforcement learning to route mixed workload of queries while accounting for interference between prefill and decode computations. 
LMDeploy~\cite{zhang2025lmdeploy} improves generation efficiency through its mixed-precision inference engine, and its workload distribution policy supports load balancing across serving backends. vLLM's Semantic Router~\cite{liu2026vllmsemanticroutersignal} provides a configurable model routing framework that includes latency-aware model selection based on normalized percentiles of observed TTFT and time-per-output-token.

Furthermore, latency estimation has been examined at varying granularities, with suitable approximations used at each stage. Vidur~\cite{MLSYS2024_b74a8de4_vidur} combines operator profiling and kernel execution time models to evaluate workloads and deployment configurations. Roofline-driven prediction~\cite{imai2024predicting_roofline_estimator} combines analytical compute and memory estimates with regression models to capture runtime effects. Block~\cite{da2025blockbalancingloadllm} combines output-length prediction with a per-instance simulation to guide load balancing across model instances. Together, these approaches demonstrate the value of latency estimation for runtime load balancing and improved resource utilization.

Joint quality--cost--latency routing has been explored by SCORE~\cite{lakha2025faster_cheaper_justasgood} and RouteBalance~\cite{da2026routebalancefusedmodelrouting}. SCORE approximates completion latency as queueing delay plus the predicted output length multiplied by a model-specific per-token processing time. RouteBalance combines learned time-per-output-token (TPOT) predictions for each model instance with predicted output lengths of sequences to estimate latencies. Our work utilizes a serving framework and batch evolution aware latency estimator. This enables us to capture the workload evolution and prefill--decode computation interference, enabling quality--cost based routing with per-query latency constraints.

\parahead{Output-length prediction and size-aware scheduling.}
Response length estimates for queries determine the decode workload and hence the memory pressure arising due to growing KV-cache during inference, making them useful for scheduling and resource management decisions. Speculative shortest-job-first scheduling~\cite{aiops2024qiu_efficient_llm_scheduling_with_proxy_model} uses a lightweight proxy model to predict response lengths for queries. Learning-to-rank scheduling~\cite{fu2024efficient_scheduling_learning_to_rank} instead focuses on predicting the relative output-length rankings to approximate the shortest-job-first ordering. TRAIL~\cite{shahout2025dontstopmenow} uses internal LLM embeddings to generate predictions for the remaining response length during execution and guide a shortest-remaining-processing-time policy with limited preemption. 

\section{Problem formulation}
\label{sec:formulation}

\parahead{System model and notation.}
We refer to each deployed LLM, together with its associated hardware allocation and serving framework, as a \emph{model instance}, indexed by $j \in \mathcal{J}$.
For each newly arriving query $q_i$, the router estimates the accuracy, cost, and time-to-first-token (TTFT) latency associated with each candidate model instance, and selects an instance $m(i)\in\mathcal{J}$ according to the desired routing objective. We consider TTFT latency as it captures the responsiveness perceived by users and interactive applications. An illustrative diagram of the system model is shown in \Cref{fig:high-level-diagram}.

\begin{figure}[h]
    \centering    \includegraphics[width=\linewidth]{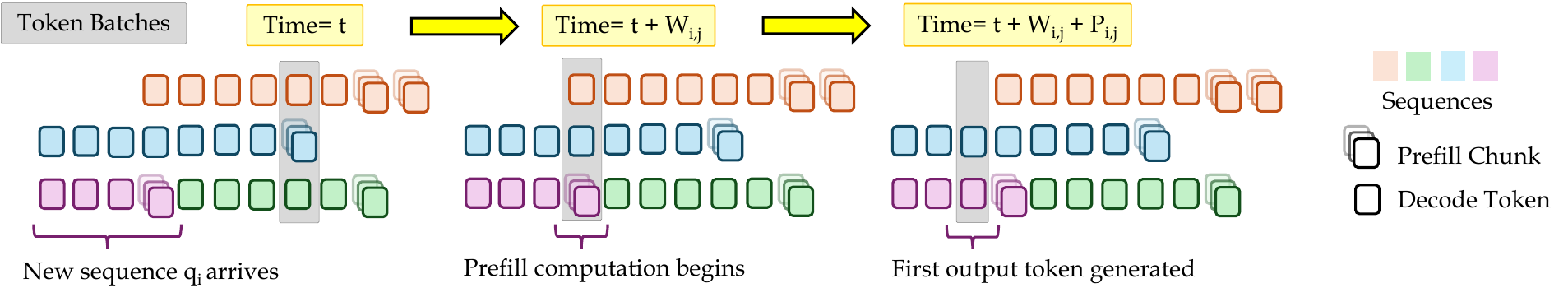}
    \caption{\small \textbf{TTFT (time-to-first-token) for autoregressive generation.} A new sequence (query) $q_i$ arrives at time $t$ to model instance $j$ (\emph{left}), its prefill computation begins at time $t\!+\!W_{i,j}$ (\emph{center}), and the first output token is generated at time $t\!+\!W_{i,j}\!+\!P_{i,j}$ (\emph{right}). New sequences are queued until the memory at the model instance is released by sequences already being generated. 
    }
    \label{fig:latency_diagram}
\end{figure}

\parahead{Generation latency.} 
The time between a sequence (query) arriving in the system and completing generation (by either generating an \texttt{<eos>} token or reaching the model's context length limit) comprises three stages: queueing delay, prefill time, and decode time. Queueing delay, or waiting time, is the idle period before a query’s prefill computation begins, during which it waits for compute and memory resources occupied by preceding queries to become available. 
We examine time-to-first-token (TTFT), which denotes the time from query arrival to first output-token generation. Thus for query $q_i$ at model instance $j$, TTFT can be represented as
\begin{align}
  L^{\text{ttft}}_{i,j}\;=\;W_{i,j}\;+\;P_{i,j}
  \label{eq:ttft-decomp}
\end{align} 
where $W_{i,j}$ represents the waiting time spent in the queue and $P_{i,j}$ is the prefill computation time, which concludes with the generation of the first output token (\Cref{subsec:background-inference}).
We illustrate the latency decomposition in \Cref{fig:latency_diagram} for continuous batching policies. Estimates ($\widehat{L}^{\text{ttft}}_{i,j}$) of TTFT for a given query $q_i$ at each model instance $j$ are required to perform latency-aware routing decisions across model instances. 

\parahead{Accuracy and cost estimators.}
Using historical evaluations of queries on language models, we train estimators for the response quality and serving cost of previously unseen queries (details presented in \Cref{subapp:acc_output_length_predictors}). For each query $q_i$ and model instance $j$, we produce an accuracy estimate $\widehat{\textrm{acc}}_{i,j}$ and a cost estimate $\widehat{\textrm{cost}}_{i,j}$. Response accuracy refers broadly to a task-dependent quality measure (such as correctness, ROUGE \cite{lin-2004-rouge-score}, or BARTScore \cite{yuan2021bartscore}). We utilize a normalized LLM-as-a-judge \cite{li-etal-2025-generation_llm-as-judge} quality score in $[0,1]$ for our analysis. We compute generation cost using the per-token pricing of each model. We define the accuracy--cost utility \cite{jitkrittum2026universal,patel2025proxrouterproximityweightedllmquery,hu2024routerbench} as:
\begin{align}
  \widehat{U}_{i,j}(\lambda)
  \;=\;
  \widehat{\textrm{acc}}_{i,j}
  -
  \lambda \widehat{\textrm{cost}}_{i,j},
  \label{eq:utility}
\end{align}
where $\lambda \ge 0$ governs the tradeoff between response quality and monetary cost.

\parahead{Routing objectives and evaluation protocols.}
In practical deployments, queries are subject to generation latency requirements based on downstream tasks. We therefore define the primary routing objective as maximizing the accuracy--cost utility $\widehat{U}_{i,j}$, while satisfying a per-query time-to-first-token (TTFT) constraint. Let $\tau_i$ denote the TTFT target for query $q_i$, and define the estimated feasible set of model instances as $\mathcal{J}(i)=\{j\in\mathcal{J}:\widehat L^{\textrm{ttft}}_{i,j}(t)\le \tau_i\}$.
The router selects the feasible model instance with the highest estimated utility (if $\mathcal{J}(i)=\phi$, select the instance with the least estimated latency):
\begin{equation}
\label{eq:constrained_routing_objective}
m(i)
\leftarrow
\argmax_{\substack{j\in\mathcal{J}:\,\, \widehat L^{\textrm{ttft}}_{i,j}(t)\le \tau_i}}
\widehat U_{i,j}(\lambda).
\end{equation}
To evaluate this constrained routing objective, we report the average utility achieved by the routed model instances on queries, assigning zero utility to queries that violate their latency constraints. We refer to this metric as \emph{OnTimeUtility}, capturing the average accuracy--cost utility over latency-constraint satisfying responses (where $N$ denotes the total number of queries evaluated):
\begin{equation}
    \textrm{OnTimeUtility}(\lambda)
    =
    \frac{1}{N}
    \sum_{i=1}^{N}
    U_{i,m(i)}(\lambda)\,
    \mathbf{1}\!\left\{
    L^{\textrm{ttft}}_{i,m(i)} \le \tau_i
    \right\},
\label{eq:ontime_utility}
\end{equation}
where $U_{i,m(i)}(\lambda)$ and $L^{\textrm{ttft}}_{i,m(i)}$ denote the realized utility and TTFT after routing, and $ \mathbf{1}\left\{ \cdot \right\}$ denotes the indicator function. 
We also characterize the underlying tradeoff between utility and latency by considering a Lagrangian relaxation of the constrained objective in \cref{eq:constrained_routing_objective}:
\begin{align}
  m(i)
  \;\leftarrow\;
  \argmax_{j\in\mathcal{J}}
  \Big(
  \widehat U_{i,j}(\lambda)
  -
  \delta\,\widehat L^{\textrm{ttft}}_{i,j}(t)
  \Big),
  \label{eq:lagrangian_routing_objective}
\end{align}
where $\delta\geq0$ controls the relative emphasis placed on latency. By varying $\delta$, we obtain a utility--latency tradeoff curve in terms of $(\mathbb{E}[U_{i,m(i)}(\lambda)],\mathbb{E}[L^{\textrm{ttft}}_{i,m(i)}])$.
This enables a characterization of the router's ability to trade off utility and latency independently of per-query service-level objectives. A better router attains higher utility at the same average latency, or equivalently achieves lower latency while preserving the same accuracy--cost utility.

\section{Estimating query latencies across model instances}
\label{sec:latency-estimation}

Routing queries across model instances under latency constraints requires accurate \emph{time-to-first-token} (TTFT) estimates.
In \Cref{subsec:sfs-latency-estimator}, we first develop the \emph{Batch Evolution Aware} (\bea) estimator which uses real-time workload information to accurately estimate TTFT.
When such information is unavailable, we also develop a queueing-theoretic analysis of autoregressive generation in \Cref{subsec:lps_queueing_delay} to estimate average latency.

\parahead{Setup.}
Let $\mathcal{Q}_j(t)$ denote the set of requests queued at model instance $j$ at time $t$, i.e., requests assigned to the instance whose prefill processing has not yet begun. Let $\mathcal{A}_j(t)$ denote the set of active
requests currently being served, either in the prefill or decode stage. We define the resident set of requests at model instance $j$ as
\[
    \mathcal{R}_j(t) \triangleq \mathcal{Q}_j(t) \cup \mathcal{A}_j(t).
\]
For each resident request $q\in\mathcal{R}_j(t)$, let
$\textrm{tok}^{\textrm{pre}}_{q,j}(t)$ and $\textrm{tok}^{\textrm{dec}}_{q,j}(t)$ denote the remaining number of prefill tokens to be processed and output tokens to be generated at time $t$, respectively. Consistent with \Cref{subsec:background-inference}, $\textrm{tok}^{\textrm{dec}}_{q,j}(t)$ includes the first output token if the prefill of $q$ has not yet completed, in which case the decode phase generates the remaining $\textrm{tok}^{\textrm{dec}}_{q,j}(t)-1$ tokens.
For a new query $q_i$ arriving at time $t$, the goal of our latency estimator is to compute the TTFT estimate $\widehat L^{\text{ttft}}_{i,j}(t)$ for each model instance $j$.

\subsection{Batch Evolution Aware (BEA) latency estimator}
\label{subsec:sfs-latency-estimator}
We propose the \bea latency estimator, which simulates how token batches evolve under the serving framework's batching and scheduling policies at each candidate model instance, using the remaining prefill workload and estimated decode workload of resident sequences. The \bea estimator accumulates the predicted processing times of successive token batches until the incoming query $q_i$ produces its first output token, resulting in the TTFT estimate $\widehat L^{\textrm{ttft}}_{i,j}(t)$. We first motivate this design by presenting a simple prefill throughput-based latency estimator and highlighting its limitations. We then elaborate how the \bea estimator incorporates predicted decode lengths together with a calibrated token-batch processing time model.

\begin{figure}[t]
    \centering
    \begin{minipage}{0.65\columnwidth}
        \centering
        \begin{subfigure}[t]{0.49\linewidth}
            \centering
            \includegraphics[width=\linewidth]{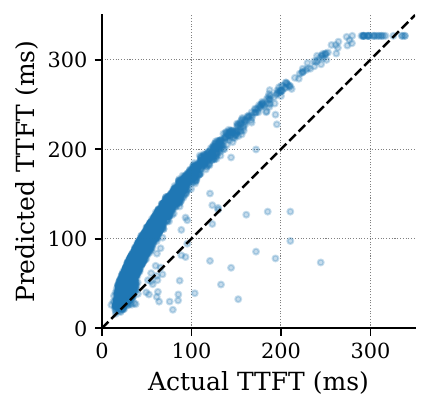}
            \caption{Prefill Throughput Estimator}
        \end{subfigure}
        \hfill
        \begin{subfigure}[t]{0.49\linewidth}
            \centering
            \includegraphics[width=\linewidth]{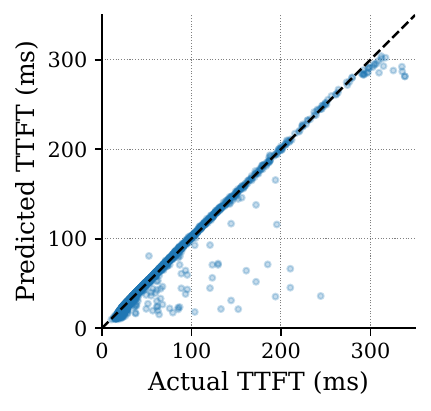}
            \caption{\bea Estimator (Ours)}
        \end{subfigure}
    \end{minipage}
    \caption{\small Comparing prefill throughput-based latency estimator \emph{(left)}
    vs. the proposed \bea estimator \emph{(right)} for Qwen3-0.6B (hosted on one H100 GPU). The prefill throughput-based latency estimator
   does not explicitly account for decode interference while the \bea estimator simulates autoregressive generation to capture it.}
   \label{fig:ttft_estimator_comparison}
\end{figure}

\parahead{Prefill throughput-based latency estimator.} 
An elementary approach to TTFT estimation for incoming queries computes the remaining prefill workload at each model instance and utilizes the prefill throughput of the model instance to predict latencies.
For each model instance $j$, we define the effective prefill throughput $\theta^{\textrm{pre}}_j$, measured in tokens per second, as the total number of prefill tokens processed across all token batches containing prefill chunks during profiling, divided by the total processing time of those batches.
At model instance $j$, the prefill throughput-based TTFT estimate (\cref{eq:ttft-decomp}) for a query $q_i$ arriving at time $t$ is
\begin{align}
\widehat L_{i,j}^{\textrm{ttft}}(t)= \underbrace{\frac{  \sum_{q\in \mathcal{R}_j(t)}  \textrm{tok}^{\textrm{pre}}_{q,j}(t)}{\theta^{\textrm{pre}}_j}}_{\text{Waiting time } \widehat W_{i,j}(t)}  \quad+ \! \underbrace{\frac{\textrm{tok}^{\textrm{pre}}_{q_i,j}(t)}{\theta^{\textrm{pre}}_j} \ ,}_{\text{Prefill and first-token time } \widehat P_{i,j}(t)}
  \label{eq:prefill-throughput-estimator}
\end{align}

where the estimated waiting time $\widehat W_{i,j}(t)$ is the sum of the remaining prefill tokens of resident queries at model instance $j$, divided by the prefill throughput $\theta^{\textrm{pre}}_j$, and $\widehat P_{i,j}(t)$ is the estimated prefill computation time for query $q_i$, which concludes with the generation of its first output token.

The prefill throughput-based latency estimator approximates the experienced TTFT latencies using the prefill workload at the model instance. This approximation aligns with prefill-prioritizing and prefill-chunk-based serving frameworks, where either prefills are computed eagerly upon query arrivals \cite{280922_orca, 10.1145/3600006.3613165_vllm}, or chunks of prefill tokens \cite{10.5555/3691938.3691945_sarathi} are interleaved with decode computations to improve hardware utilization. 
However, this estimator does not explicitly account for decode-workload interference and its impact on memory and compute utilization at model instances. The calibrated $\theta^{\textrm{pre}}_j$ only indirectly captures the effect of interleaved decode computations.
Consequently, as shown in \Cref{fig:ttft_estimator_comparison} \emph{(left plot)}, the prefill throughput-based latency estimator suffers bias when the evaluation-time prefill-decode workload distribution differs from the throughput profiling workload (note that the few outliers in both plots in \Cref{fig:ttft_estimator_comparison} arise from CPU and kernel overheads).

\parahead{BEA estimator.}
To improve beyond the prefill throughput-based latency estimator described above, we propose the \bea estimator, which jointly considers the prefill and decode workloads at model instances. It considers the remaining prefill workload $\{\textrm{tok}^{\textrm{pre}}_{q,j}(t)\}_{q\in\mathcal{R}_j(t)}$ for queries in the current resident set $\mathcal{R}_j(t)$ at model instance $j$, and estimates the remaining number of output tokens $\{\widehat{\textrm{tok}}^{\textrm{dec}}_{q,j}(t)\}_{q\in\mathcal{R}_j(t)}$ by using an output length estimator (\Cref{subapp:acc_output_length_predictors}). The workload of the resident query set is augmented with prefill token length, and output length estimates $\left(\textrm{tok}^{\textrm{pre}}_{q_i,j}(t),\widehat{\textrm{tok}}^{\textrm{dec}}_{q_i,j}(t)\right)$ for the new query $q_i$. Subsequently, the \bea estimator simulates the serving framework's batching and scheduling policies until the first output token for sequence $q_i$ is generated.
Concretely, when estimating TTFT generation latency at model instance $j$ for the arrival of query $q_i$ at time $t$, the \bea estimator forms the augmented workload
\begin{equation}
    \mathcal{X}_{i,j}(t)
    =
    \left\{
    \left(
    \textrm{tok}^{\textrm{pre}}_{q,j}(t),
    \widehat{\textrm{tok}}^{\textrm{dec}}_{q,j}(t)
    \right)
    \;:\;
    q\in\mathcal{R}_j(t)
    \right\}
    \cup
    \left\{
    \left(
    \textrm{tok}^{\textrm{pre}}_{q_i,j}(t),
    \widehat{\textrm{tok}}^{\textrm{dec}}_{q_i,j}(t)
    \right)
    \right\}.
\end{equation}
Using $\mathcal{X}_{i,j}(t)$, the estimator then deterministically simulates the serving framework, resulting in consecutive token batches for autoregressive generation. Here, a \emph{token batch} refers to the set of token computations to be performed, consisting of at most one decode token per active decode sequence and possibly a prefill chunk for one or more sequences \cite{10.5555/3691938.3691945_sarathi}.

Let $\mathcal{B}^{(\ell)}_j(t)$ denote the $\ell^\textrm{th}$ token batch from time $t$ at model instance $j$. 
For each sequence $q$ being processed in $\mathcal{B}^{(\ell)}_j(t)$, let $\textrm{tok}^{\textrm{pre}}_{q,j}(t,\ell)$ be the number of prefill tokens to be processed, and $\textrm{tok}^{\textrm{dec}}_{q,j}(t,\ell)$ be the number of decode tokens to be generated in batch $\ell$.
In the standard autoregressive generation setting, each sequence produces at most one decode token per batch, so
$\textrm{tok}^{\textrm{dec}}_{q,j}(t,\ell)\in\{0,1\}$.
The prefill phase concludes by generating the first output token (\Cref{subsec:background-inference}), so decode tokens refer to the second output token onwards.
Accordingly, the TTFT of $q_i$ is determined by the earliest batch in which all of its remaining prefill tokens $\textrm{tok}^{\textrm{pre}}_{q_i,j}(t)$ have been processed:
\begin{equation}
\label{eq:first_output_token_batch}
    \ell^{\textrm{ttft}}_j(t,i)
    \triangleq
    \min \left\{
    \ell:
    \sum_{\ell'=1}^{\ell}
    \textrm{tok}^{\textrm{pre}}_{q_i,j}(t,\ell')
    =
    \textrm{tok}^{\textrm{pre}}_{q_i,j}(t)
    \right\}.
\end{equation}
\Cref{fig:latency_diagram} illustrates the model instance state and token-batch composition when a new sequence $q_i$ arrives (\emph{left}), when its prefill computation begins (\emph{center}), and when its first output token is generated (\emph{right}), marking the TTFT for $q_i$.
The \bea estimator obtains the TTFT estimate by summing the predicted processing times of the simulated token batches up to and including the batch that generates the first output token for $q_i$:
\begin{equation}
\label{eq:sfs_ttft_sum}
    \widehat{L}^{\textrm{ttft}}_{i,j}(t)
    \triangleq
    \sum_{\ell=1}^{\ell^{\textrm{ttft}}_j(t,i)}
    \widehat{T}^{(\ell)}_j(t),
\end{equation}
where $\widehat{T}^{(\ell)}_j(t)$ (defined next) denotes the time required to process the token batch $\mathcal{B}^{(\ell)}_j(t)$.
Unlike the prefill throughput-based estimate in \cref{eq:prefill-throughput-estimator}, the \bea estimator does not
separately approximate waiting and prefill (including first output token) computation using model instance throughputs, instead capturing these components directly via the simulated token batches. \Cref{fig:ttft_estimator_comparison} demonstrates that the \bea estimator generates better estimates of TTFT latency compared to the prefill throughput-based latency estimator ($5\%$ vs $85\%$ mean absolute percentage error) by jointly considering prefill and decode workloads and modeling autoregressive generation at model instances.

\parahead{Estimating the token-batch processing time $\widehat{T}^{(\ell)}_j(t)$.}
Processing a token batch at a model instance primarily involves KV-cache memory reads, MLP and attention computations. Modeling each low-level kernel and memory transfer separately is tedious and difficult to generalize across serving workloads. We therefore design a higher-level estimator based on token batch composition.
For each sequence $q\in\mathcal{B}^{(\ell)}_j(t)$, let
$c_{q,j}(t,\ell)$ denote the context length of sequence $q$ before token batch $\ell$ is processed. We estimate the processing
time of token batch $\mathcal{B}^{(\ell)}_j(t)$ as
\begin{align}
\widehat T^{(\ell)}_j(t)
=
\beta_{0,j}
&+
\beta_{1,j}
\sum_{q\in \mathcal{B}^{(\ell)}_j(t)}
\left(
\textrm{tok}^{\textrm{pre}}_{q,j}(t,\ell)
+
\textrm{tok}^{\textrm{dec}}_{q,j}(t,\ell)
\right)
+
\beta_{2,j}
\sum_{q\in \mathcal{B}^{(\ell)}_j(t)}
c_{q,j}(t,\ell)\,
\textrm{tok}^{\textrm{dec}}_{q,j}(t,\ell)
\nonumber\\
&+
\beta_{3,j}
\sum_{q\in \mathcal{B}^{(\ell)}_j(t)}
\left(
\textrm{tok}^{\textrm{pre}}_{q,j}(t,\ell)c_{q,j}(t,\ell)
+
\frac{
\textrm{tok}^{\textrm{pre}}_{q,j}(t,\ell)
\left(\textrm{tok}^{\textrm{pre}}_{q,j}(t,\ell)+1\right)
}{2}
\right).
\label{eq:batch_processing_time_estimator}
\end{align}

\begin{wrapfigure}{r}{0.42\columnwidth}
    \centering
    % \vspace{1pt}
    \includegraphics[width=0.84\linewidth]{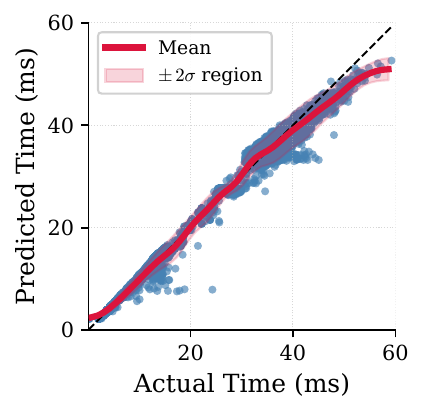}
    \caption{\small \textbf{Token-batch processing time in the \bea estimator.} Batch processing times predicted by \cref{eq:batch_processing_time_estimator} closely match observed batch times, with $\approx\!4\%$ mean absolute percentage error across token-batch compositions (Qwen3-0.6B on one H100 GPU).}
    \label{fig:batch_processing_time}
    \vspace{-5pt}
\end{wrapfigure}
The intercept term ($\beta_{0,j}$) captures fixed per-batch overheads. The first workload term (with coeff. $\beta_{1,j}$) captures token-wise dense-layer computation, which scales linearly with the number of tokens in a batch. The second term (with coeff. $\beta_{2,j}$) accounts for the attention computation and KV-cache reads for each decode sequence, which scales linearly with increasing context. 
The third term (with coeff. $\beta_{3,j}$) models prefill attention cost: each prefill token attends both to the existing context and to the preceding tokens within the same prefill chunk, yielding a contribution that grows with the product of the prefill chunk size and prior context length, plus the quadratic attention cost within the chunk.

The coefficients $(\beta_{0,j},\beta_{1,j},\beta_{2,j},\beta_{3,j})$ are calibrated for each model instance using observed token-batch processing times. \Cref{fig:batch_processing_time} shows the estimated token-batch processing time from \cref{eq:batch_processing_time_estimator} against the observed batch processing time, demonstrating high accuracy with $\approx\!4\%$ mean absolute percentage error.

\subsection{Analytical model for autoregressive generation}
\label{subsec:lps_queueing_delay}

\begin{wrapfigure}[18]{r}{0.43\columnwidth}
\vspace*{-1em}
    \centering
    \includegraphics[width=\linewidth]{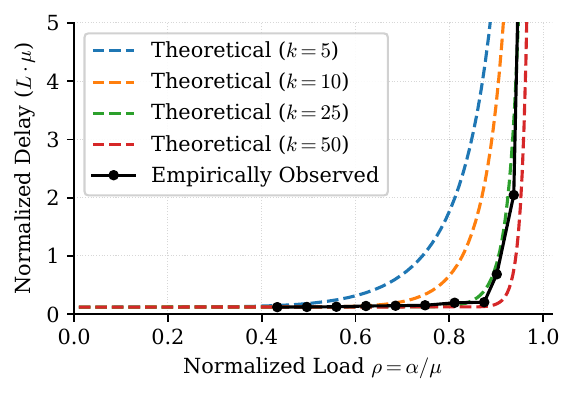}
    \caption{Measured delay for Qwen3-0.6B on a single H100 GPU across arrival workloads, compared with the analytical predictions of the limited processor sharing model (\cref{eq:lps_mean_wait,eq:lps_ttft}). The measured latencies closely follow the predicted latencies with concurrency level $k\!=\!25$, consistent with the observed average of $25.7$ active sequences during generation.}
    \label{fig:theoretical_vs_empirical_ttft_queueing_lps}
\end{wrapfigure}

The \bea estimator relies on real-time workload information at model instances, which may not always be available to the router. For such cases, we develop an analytical model of autoregressive 
\begin{wrapfigure}[15]{r}{0.42\columnwidth}
\vspace*{-1.0em}
    \centering
    \includegraphics[width=\linewidth]{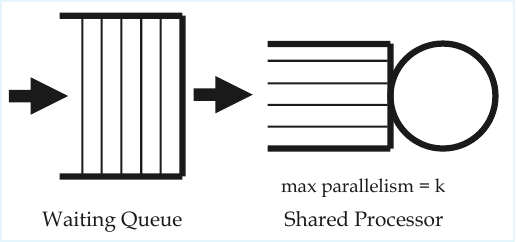}
    \caption{\small \textbf{Limited processor sharing (LPS) model} \cite{zhang2009limited_processor_sharing}. Up to $k$ jobs (actively decoded queries) share the processor (the compute of a model instance) equally. A new job begins service immediately if fewer than $k$ jobs are present, and otherwise waits in a first-come-first-served (FCFS) queue until an active job completes.}
    \label{fig:limited_processor_sharing_diagram}
\end{wrapfigure}
generation that estimates the expected TTFT from time-average arrival workloads and serving capacities of model instances. We model each instance through a limited processor sharing (LPS) \cite{zhang2009limited_processor_sharing} abstraction, derive its expected queueing delay using queueing-theoretic results, and empirically corroborate the resulting estimates in \Cref{fig:theoretical_vs_empirical_ttft_queueing_lps}.

\parahead{LPS abstraction of a model instance.}
As illustrated in \Cref{fig:limited_processor_sharing_diagram}, the LPS abstraction of a model instance $j\in\mathcal{J}$ captures two properties of modern serving frameworks. First, continuous batching processes multiple sequences concurrently \cite{280922_orca}, but the number of concurrently active sequences is limited by KV-cache memory and hardware constraints \cite{10.1145/3600006.3613165_vllm}. We capture this limit by a concurrency cap $k_j$, with excess queries waiting in an FCFS queue. Second, compute capacity is shared across actively decoded sequences, so each sequence progresses faster under light load and slower under heavy load \cite{10.5555/3691938.3691945_sarathi,10.1145/3600006.3613165_vllm}. We capture this sharing by processor sharing with an aggregate service rate of $\mu_j$ queries per second.

Formally, let $|\mathcal{R}_j(t)|$ denote the number of queries resident at instance $j$ (queued or active), and $|\mathcal{A}_j(t)|$ the number of active queries. Under LPS, the first $|\mathcal{A}_j(t)|\!=\!\min\{|\mathcal{R}_j(t)|,k_j\}$ queries are in service, each at instantaneous rate
${\mu_j}/{\min\{|\mathcal{R}_j(t)|,k_j\}},$
and a query remains active until its response is complete. Both parameters can be measured directly at each instance: $\mu_j$ is its throughput in queries per second, and $k_j$ is the number of concurrently decoding sequences at saturation.

\parahead{Assumptions.} (1)~\textbf{Poisson arrivals:} query arrivals form a Poisson process of rate $\alpha_j$ queries per second.\footnote{\Cref{app:lps_general_approx} relaxes assumptions (1) and (2) to general arrival processes and service requirement distributions.} (2)~\textbf{Memoryless service requirement:} each query has an independent and identically distributed exponential service requirement with mean $1/\mu_j$, i.e., the time to complete the query when served at the full capacity of the model instance. Alternatively, generation work for queries follows the memoryless distribution \cite{10.1145/3721146.3721962_geometric_decode_distribution}. (3)~\textbf{Limited parallelism:} at most $k_j$ queries can be served simultaneously, and the remaining queries wait in an FCFS queue. (4)~\textbf{Saturated aggregate capacity:} The model instance shares its capacity equally amongst active queries with a constant aggregate service rate $\mu_j$. (5)~\textbf{Stability condition:} the arrival rate of queries is strictly below the service capacity, i.e., the utilization $\rho_j=\alpha_j/\mu_j<1$.

\parahead{State process and stationary distribution.}
Queries arrive according to a Poisson process of rate $\alpha_j$, and each arrival increases the resident set size $|\mathcal{R}_j(t)|$ by one.
For $|\mathcal{R}_j(t)|$ resident queries in the system, with $|\mathcal{A}_j(t)|$ active queries, the total service capacity $\mu_j$ is shared equally among those active queries. 
Thus, the process $\{|\mathcal{R}_j(t)|\}_{t\ge 0}$ behaves as a birth--death chain with upward transition rate $\alpha_j$ and downward transition rate $\mu_j$ for every nonzero state. 
Hence, the stationary distribution of $\{|\mathcal{R}_j(t)|\}_{t\geq 0}$ coincides with that of an $M/M/1$ queue:
\[
\pi_n \triangleq \Pr(|\mathcal{R}_j|=n) = (1-\rho_j)\rho_j^n,\qquad n=0,1,2,\ldots.
\]

\parahead{Queueing delay of an arriving query.}
Let $W_{i,j}$ denote the queueing delay, i.e., the time from arrival until the query $q_i$ first enters the active set at model instance $j$. By PASTA (Poisson Arrivals See Time Averages) \cite{c62d9850-ac5b-3ff7-94cf-2d2ab3b8d389_pasta}, an arrival sees state $|\mathcal{R}_j(t)|=n$ with probability $\pi_n$.
A new query begins service immediately if the resident set size ($n$) is smaller than the concurrency cap ($k_j$), hence
\[
\mathbb{E}[W_{i,j}\bigm| |\mathcal{R}_j(t)|=n]=0,\qquad n<k_j.
\]
Alternatively, if the resident set completely occupies the concurrency level ($n\ge k_j$), then $k_j$ jobs are already active, and $n-k_j$ jobs are queued. Since the new query $q_i$ joins the FCFS queue, it must wait for $n-k_j+1$ service completions before entering service. Throughout the waiting period for query $q_i$, the server always has $k_j$ active jobs, and the existing jobs complete as a Poisson process of rate $\mu_j$. Later arrivals after query $q_i$ join the FCFS queue in order, and thus do not change the number of completions required for the admission of query $q_i$ at the server. Hence
\[
\mathbb{E}\left[W_{i,j}\bigm| |\mathcal{R}_j(t)|=n\right]=\frac{n-k_j+1}{\mu_j},\qquad n\ge k_j.
\]
Averaging over the stationary state observed by query arrivals results in
\begin{align*}
\mathbb{E}[W_{i,j}]
&= \sum_{n=0}^{\infty}\pi_n\,\mathbb{E}[W_{i,j}\bigm| |\mathcal{R}_j(t)|=n] = \sum_{n=k_j}^{\infty}(1-\rho_j)\rho_j^n \frac{n-k_j+1}{\mu_j} \\
&= \frac{(1-\rho_j)\rho_j^{k_j}}{\mu_j}\cdot \frac{1}{(1-\rho_j)^2} = \frac{\rho_j^{k_j}}{\mu_j(1-\rho_j)}
= \frac{(\alpha_j/\mu_j)^{k_j}}{\mu_j-\alpha_j}.
\end{align*}
Thus, the expected queueing delay estimator for query $q_i$ at model instance $j$ is,
\begin{equation}
\label{eq:lps_mean_wait}
\boxed{
    \widehat W^{\textrm{avg}}_{i,j} = \mathbb{E}[W_{i,j}] = \frac{\rho_j^{k_j}}{\mu_j-\alpha_j} =  \frac{(\alpha_j/\mu_j)^{k_j}}{\mu_j-\alpha_j}.
}
\end{equation}
For $k_j=1$, LPS reduces to the ordinary $M/M/1$, yielding $\mathbb{E}[W_{i,j}]=\rho_j/(\mu_j-\alpha_j)$, the  FCFS waiting time for $M/M/1$ queues. The waiting time reduces geometrically as $k_j$ increases, and vanishes as $k_j\!\to\!\infty$, implying immediate admission under the ordinary processor sharing scheme (although each sequence has infinitesimal serving rate).

\parahead{Time-to-first-token latency estimation.} Utilizing the analytically estimated wait time in \cref{eq:lps_mean_wait}, we estimate the TTFT for a given sequence as the sum of its queueing delay and the prefill computation time (\cref{eq:ttft-decomp}). We approximate the prefill computation time for the given query through the $j^\textrm{th}$ model instance's average prefill throughput $\theta^\textrm{pre}_j$ and the number of prefill tokens $\textrm{tok}^{\textrm{pre}}_{q_i,j}$ for the incoming query $q_i$. Consequently, the 
average case TTFT estimator is 
\begin{minipage}{0.53\columnwidth}
\begin{equation}
\widehat L^{\textrm{avg}}_{i,j}
=
\widehat W^{\textrm{avg}}_{i,j}
+
\frac{\textrm{tok}^{\textrm{pre}}_{q_i,j}}{\theta^{\textrm{pre}}_j}.
\label{eq:lps_ttft}
\end{equation}
\end{minipage}

\Cref{fig:theoretical_vs_empirical_ttft_queueing_lps} compares the analytically predicted latencies (\cref{eq:lps_mean_wait,eq:lps_ttft}) under the LPS model with the empirically observed latencies under real serving systems. 
In the chosen setup with Qwen3-0.6B model served on a single H100 GPU, the measured average parallelism for sequence generation is $\approx\!25.7$ active sequences. 
The LPS abstraction for autoregressive generation captures the qualitative trend in latencies as the query workload increases, and the measured latencies also closely match the predicted values for a concurrency level of $k\!=\!25$ in the analytical model.  
The qualitative and quantitative agreement between the estimated and observed latencies indicates that the LPS abstraction provides an accurate model of the load-dependent latency estimation arising during autoregressive generation in modern LLM serving systems. 
In \Cref{app:lps_general_approx}, we further extend the LPS waiting time to general arrival processes and output-length distributions.

\section{Experiments}
\label{sec:experiments}
We now examine our latency aware routing approach and compare with baselines. We describe the experimental setup, examined baselines, and evaluation protocol in \Cref{subsec:experimental_setup}. We then evaluate the routing policies through two complementary directions, first focusing on performance as the query workload increases (\Cref{subsec:arrival_rate_experiments}), and secondly assessing the performance as query latency constraints vary (\Cref{subsec:varying_latency_constraints}). 
Finally, we test robustness to bursty arrivals, measure the estimator's overhead and conduct additional analyses in \Cref{subsec:additional_experiments}.

\subsection{Experimental setup}
\label{subsec:experimental_setup}

\begin{wrapfigure}[17]{r}{0.5\columnwidth}
    \centering
    \vspace{-10pt}
    \includegraphics[width=0.93\linewidth]{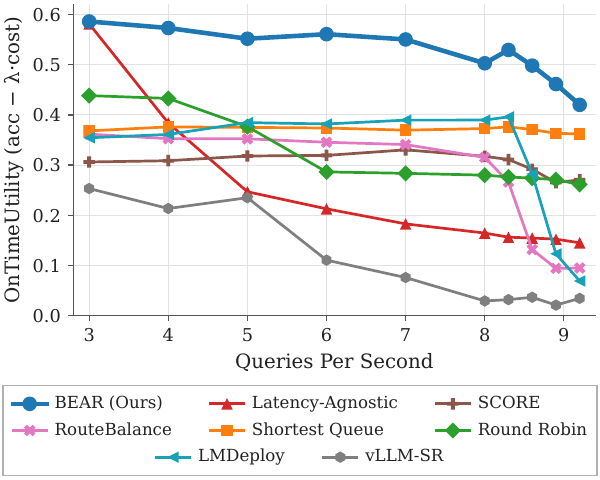}
    \caption{\small \textbf{OnTimeUtility across varying arrival rate (in qps).} \bear achieves 30.5\% higher OnTimeUtility on average than the best baseline at each arrival rate.}
    \label{fig:ontimeutility_vs_qps}
    % \vspace{-10pt}
\end{wrapfigure}

We evaluate routing on queries sampled from four tasks:
\textbf{Alpaca}~\cite{alpaca_dataset}, \textbf{HotpotQA}~\cite{yang2018hotpotqa}, \textbf{GovReport-Summarization}~\cite{huang-etal-2021-efficient_govreport_summarization}, and \textbf{WritingPrompts}~\cite{fan-etal-2018-hierarchical_writing_prompts_dataset}.
These tasks respectively assess instruction following, reasoning over supporting documents, long-document summarization, and creative generation for language models. The average prompt and response lengths of tasks belong to the (short, short), (long, short), (long, long), and (short, long) regimes respectively, evaluating the router's latency estimator to different combinations of prefill and decode workloads.
We sample $2.5$K queries from each task to estimate per-instance prefill throughput, calibrate the token-batch processing time estimators, and train output length and response quality predictors. Our primary experiments simulate query arrivals from Poisson Process (\Cref{fig:ontimeutility_vs_qps,fig:utility_vs_latency}), and we additionally evaluate correlated and bursty arrivals based on a Markov-Modulated Poisson process (\Cref{fig:query_arrival_distribution_ablation}).

Our language model pool comprises of \textbf{Qwen3-0.6B}, \textbf{Qwen3-8B}, and \textbf{Qwen3-32B} from the Qwen3 family \cite{yang2025qwen3technicalreport}, covering a broad range of sizes and capabilities. We deploy Qwen3-0.6B and Qwen3-8B on one H100 GPU \cite{nvidia-h100-pcie-product-brief-2022} each, and Qwen3-32B on two H100 GPUs (using tensor parallelism), characterizing three model instances. 
We use the vLLM \cite{10.1145/3600006.3613165_vllm} as the backbone serving framework for our experiments; however, our presented latency estimators can be easily extended to other serving frameworks as well.
We evaluate the responses generated by models on queries via LLM-as-a-judge \cite{li-etal-2025-generation_llm-as-judge} using Gemini 3.1 Pro Preview and Gemini 2.5 Flash \cite{geminiteam2025geminifamilyhighlycapable} (evaluation prompt provided in \Cref{app:llm-as-judge}). We use LightGBM \cite{NIPS2017_6449f44a_lightgbm} regressors over prompt features and token embeddings to generate accuracy predictions and decode length predictions (inference cost is estimated from token pricing through prefill and predicted decode lengths). 
We discuss further experimental details in \Cref{app:query_model_details}.

\parahead{Baselines.}
We compare our approach against other accuracy--cost--latency based routers and job scheduling and load balancing policies implemented within our experimental setup. 
\emph{RouteBalance} \cite{da2026routebalancefusedmodelrouting} combines predicted response quality with cost and latency scores normalized across candidate instances.
\emph{SCORE} \cite{lakha2025faster_cheaper_justasgood} applies a tunable coefficient to balance predicted response quality against weighted cost and latency penalties. 
\emph{Latency-Agnostic} routing policy maximizes predicted accuracy--cost utility without accounting for latency.
\emph{Round Robin} policy assigns queries cyclically across all model instances.
\emph{Shortest Queue} routing policy routes incoming query to the instance with the fewest resident requests, including queued and active requests.
\emph{LMDeploy-Proxy}~\cite{zhang2025lmdeploy} selects the instance with the smallest throughput-normalized request count, computed as the number of unfinished requests divided by measured throughput in requests per second.
\emph{vLLM Semantic Router}'s~\cite{liu2026vllmsemanticroutersignal} latency minimization approach routes queries to model instances based on normalized percentiles of observed latencies.

\parahead{Evaluation protocol.} We examine latency-aware routing from two complementary perspectives. First, we vary the offered query load under the latency-constrained routing objective in \cref{eq:constrained_routing_objective}. 
We report the \emph{OnTimeUtility} (\cref{eq:ontime_utility}), i.e. the average realized accuracy--cost utility over all queries, and nominally assigning zero contribution to responses that violate their TTFT targets.
Second, we vary the latency penalty parameter $\delta$ under the Lagrangian relaxation of latency-constrained routing~(\cref{eq:lagrangian_routing_objective}). For both these scenarios, we keep the cost coefficient $\lambda$ fixed (at $5\times10^{-4}$) and defer sweeps over $\lambda$ to \Cref{app:additional_results}.
We further examine robustness of routing performance, discuss latency estimation overhead, and present additional analysis.

\subsection{Varying arrival rate of queries}
\label{subsec:arrival_rate_experiments}
As the arrival rate of queries increases, the model instances become saturated and the feasible set of instances that satisfy queries' latency constraints shrinks. In such scenarios, a better router will yield a higher \emph{OnTimeUtility} value (\cref{eq:ontime_utility}) for latency-constrained routing (\cref{eq:constrained_routing_objective}) across a wide range of query workloads. 
\Cref{fig:ontimeutility_vs_qps} shows that \bear, using our proposed \bea latency estimator, achieves the highest \emph{OnTimeUtility} across all query workloads, yielding an average  \textbf{$\mathbf{63\%}$ gain} in OnTimeUtility over the latency aware routing policies (RouteBalance \cite{da2026routebalancefusedmodelrouting}, SCORE \cite{lakha2025faster_cheaper_justasgood}), with a maximum gain of \textbf{$\mathbf{74\%}$} at $8.9$ queries per second. 
Relative to the best baseline at each arrival rate, \bear improves OnTimeUtility by \textbf{$\mathbf{30.5\%}$} on average.

\parahead{Routing composition across model instances.}
\Cref{fig:routing_composition} illustrates how the routing fraction across model instances changes based on the arrival rate of queries.
At higher arrival rates, our latency-aware policy routes an increasing fraction of queries to the smaller and faster Qwen3-0.6B model, resulting in higher TTFT target satisfaction, while retaining substantial use of larger models when latency constraints permit.
Latency-Agnostic and Round Robin policies maintain approximately the same allocation fractions across arrival rates, leading to frequent SLO violations as chosen model instances become more congested. Join the shortest queue routing policy maintains high TTFT SLO attainment percentages, but results in a suboptimal use of larger and more capable model instances. 
These patterns illustrate how jointly optimizing latency and accuracy--cost utility across heterogeneous model instances enables more effective resource utilization and improves overall routing performance.

\vspace*{-0.5em}
\subsection{Varying latency constraints for queries}
\label{subsec:varying_latency_constraints}

\begin{wrapfigure}{r}{0.5\columnwidth}
    \centering
    \vspace{-2em}
    \includegraphics[width=0.93\linewidth]{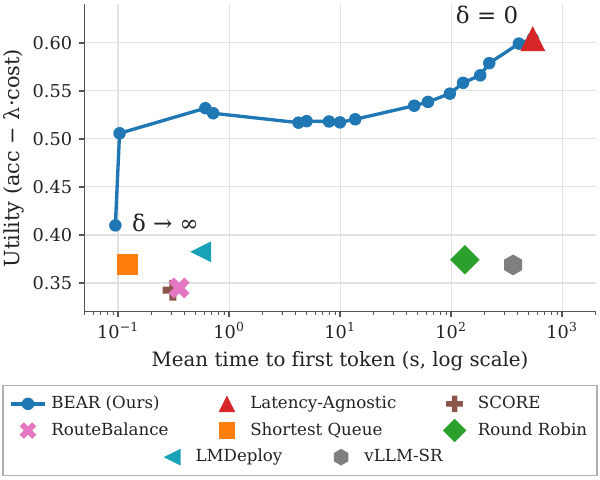}
    \caption{Varying $\delta$ in \cref{eq:lagrangian_routing_objective}. Latency-aware routing provides a controllable utility--latency tradeoff. 
    }
    \label{fig:utility_vs_latency}
    \vspace{-1em}
\end{wrapfigure}

To analyze the trade-off between utility and latency, we vary $\delta$ values in \cref{eq:lagrangian_routing_objective} and obtain average utility and latency values for queries (following the Poisson arrival process with rate $=\!8$ qps). For RouteBalance \cite{da2026routebalancefusedmodelrouting} and SCORE \cite{lakha2025faster_cheaper_justasgood} baselines, we present the OnTimeUtility maximizing configuration. 
\Cref{fig:utility_vs_latency} shows that our proposed approach consistently improves the utility--latency performance over the baselines, and importantly, provides a controllable tradeoff which can be set by choosing the corresponding $\delta$ value. Setting $\delta\!=\!0$ recovers the \emph{Latency-Agnostic} routing policy, and choosing $\delta\!\to\!\infty$ purely minimizes inference TTFT latency. A better routing policy exhibits a higher accuracy--cost utility at comparable TTFT, or equivalently demonstrates a lower TTFT at the same observed utility value. 

\subsection{Additional empirical evaluation}
\label{subsec:additional_experiments}
\begin{figure}[t]
    \centering
    \begin{minipage}[t]{0.53\linewidth}
        \centering
        \includegraphics[width=0.95\linewidth]{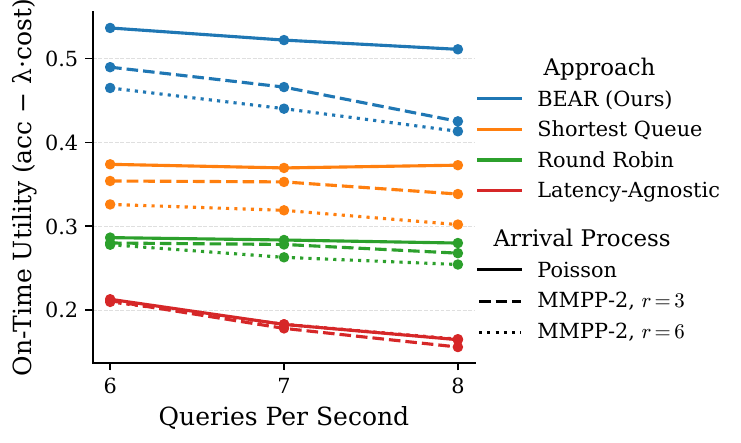}
        \caption{\small OnTimeUtility under two-state Markov-modulated Poisson arrivals with high-to-low arrival-rate ratios $r=3,6$, evaluated across average arrival rates.}
        \label{fig:query_arrival_distribution_ablation}
    \end{minipage}
    \hfill
    \begin{minipage}[t]{0.43\linewidth}
        \centering
        \includegraphics[width=0.95\linewidth]{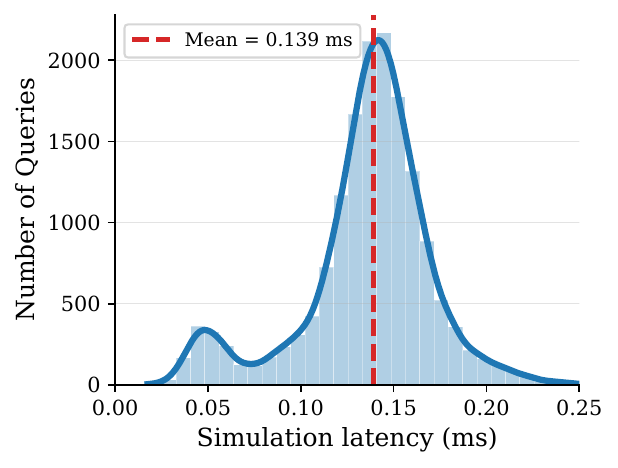}
        \caption{\small Wall-clock execution time of the \bea latency estimator. The mean simulation time for latency estimation is approximately $10^{-4}$\,s.}
        \label{fig:simulation_latency}
    \end{minipage}
    \vspace*{-1em}
\end{figure}

\parahead{Evaluation across arrival processes.}
 In practical deployments, query arrivals are often correlated and bursty, which may not be captured well by the standard Poisson arrival process. We extend our analysis to query arrivals based on a two-state Markov Modulated Poisson Process (MMPP-2) \cite{fischer1993mmpp} (\Cref{app:query_arrival_distribution_ablation}), where the low-state denotes standard query arrivals (rate $\alpha_L$) and a high-state indicates a duration of heavy workload (with arrival rate $\alpha_H$).
\Cref{fig:query_arrival_distribution_ablation} shows that our approach maintains a consistently higher OnTimeUtility across arrival rates and processes,  demonstrating the benefit of adapting model selection to real-time serving conditions during fluctuations in demand.

\parahead{Estimator overhead.}
Our proposed latency estimator simulates batching policies from the current workload state (including the newly arrived query) and halts at the token batch corresponding to first token generation for this new sequence. This avoids full simulation till completion of all resident requests at the model instance. 
\Cref{fig:simulation_latency} depicts an average latency estimation \textbf{time overhead of $\mathbf{\approx\!10^{-4}}$s}, which is negligible relative to query generation TTFTs (usually in seconds \cite{10.1145/3600006.3613165_vllm}). Further implementation details of our latency estimator are provided in \Cref{app:batching_simulation_implementation}.

\parahead{Additional ablations and sensitivity analyses.} We present additional robustness and sensitivity analyses in \Cref{app:additional_results}.  
We first examine the impact of communication delay between model instances and the router for exchanging real-time workload information (\Cref{fig:snapshot_staleness}), and then study the sensitivity of the response quality scores to the choice of LLM-as-judge (\Cref{fig:judge_score_distributions}). 
We further investigate the improvement in routing performance attributed to the latency estimator in \Cref{fig:prefill_throughput_ablation}. 
We also examine sensitivity to the accuracy--cost coefficient (\Cref{fig:lambda_sweep_results}) and present TTFT SLO attainment across arrival rates (\Cref{fig:slo_attainment_vs_qps}). 
Lastly, in \Cref{fig:latency_estimation_comparison}, we comprehensively analyze the performance of the underlying latency estimators used by various approaches to better contextualize the gains of \bear over the baselines observed in \Cref{fig:ontimeutility_vs_qps}.

\section{Concluding Remarks}
\label{sec:conclusion}
Most existing language query routers optimize the accuracy--cost trade-off, but do not account for latencies experienced by queries. We proposed \bear, a latency-aware router that jointly optimizes accuracy, cost, and time-to-first-token (TTFT) latency. The core component of our approach is the lightweight Batch Evolution Aware (\bea) latency estimator that predicts TTFT by simulating prefill and decode token batches under the serving framework's policies. Promising future directions include extension to general scheduling policies beyond First-Come, First-Served (FCFS), and accounting for KV-cache affinity in multi-turn response settings.

\bibliographystyle{ACM-Reference-Format}
\bibliography{references}

\clearpage
\appendix

\section{Query and language model details}
\label{app:query_model_details}

\subsection{Model inference costs and average performance}
\label{subapp:model_inference_costs}

\Cref{tab:qwen3_token_costs} reports the tokenwise input and output costs for the language models used in our analysis. \Cref{tab:qwen3_dataset_scores} presents the average performance of models across the tasks under consideration, evaluated using the LLM-as-a-judge scheme \cite{li-etal-2025-generation_llm-as-judge}.

\begin{table}[h]
\centering
\caption{Pricing for Qwen3 models (non-thinking mode) in USD per Million tokens on Alibaba Cloud Model Studio \cite{alibaba_model_studio_pricing_2026}. 
}
\label{tab:qwen3_token_costs}
\begin{tabular}{lcc}
\toprule
Model & Prompt & Response \\
\midrule
Qwen3-0.6B & 0.044 & 0.173 \\
Qwen3-8B   & 0.072 & 0.287 \\
Qwen3-32B  & 0.287 & 0.640 \\
\bottomrule
\end{tabular}
\end{table}

\begin{table}[h]
\centering
\caption{Per-query task average scores (using LLM-as-judge \cite{li-etal-2025-generation_llm-as-judge}) for Qwen3-0.6B/8B/32B models \cite{yang2025qwen3technicalreport} across Alpaca \cite{alpaca_dataset}, Govreport-Summarization \cite{huang-etal-2021-efficient_govreport_summarization}, HotpotQA \cite{yang2018hotpotqa}, WritingPrompts \cite{fan-etal-2018-hierarchical_writing_prompts_dataset} tasks.}
\label{tab:qwen3_dataset_scores}
\begin{tabular}{lccc}
\toprule
Query Task & Qwen3-0.6B & Qwen3-8B & Qwen3-32B \\
\midrule
Alpaca                  & 53.49\% & 82.67\% & 88.85\% \\
GovReport-Summarization & 29.02\% & 86.62\% & 95.93\% \\
HotpotQA (distractor)   & 40.98\% & 88.28\% & 92.69\% \\
WritingPrompts          & 17.73\% & 57.47\% & 80.69\% \\
\midrule
Aggregate               & 35.31\% & 78.76\% & 89.54\% \\
\bottomrule
\end{tabular}
\end{table}

\subsection{TTFT latency assignment for queries}
\label{subapp:ttft-slo-details}

For evaluations under the hard-constrained routing setting (\cref{eq:constrained_routing_objective}), each request $i$ is assigned a TTFT target $\tau_i$. We choose realistic and practically achievable TTFT constraints based on our experimental setup, including language models and hardware. The chosen constraints are a simple linear function of the prompt length $p_i$ and include a small amount of random variation. Specifically, we sample
$u_i\sim\textrm{Unif}(1\textrm{ms},5\textrm{ms})$ and $v_i\sim\textrm{Unif}(0.98,1.02)$, and set
\[
\tau_i
=
\min\!\Bigl\{
1120\textrm{ms},
\max\!\bigl\{
150\textrm{ms},
u_i + \bigl(155 + 3.5\times10^{-3}p_i\bigr)v_i
\bigr\}
\Bigr\}.
\]
Under this parameterization, the expected unclipped target is approximately
\[
\mathbb{E}[\tau_i] \approx 158 + 3.5\times10^{-3}p_i \text{ ms}.
\]
All TTFT targets are generated using a seeded pseudorandom number generator for reproducibility.

\newpage
\subsection{Prompt (prefill) and response (decode) length distributions}
\label{subapp:token_length_distributions}
\begin{figure}[h]
    \centering
    \includegraphics[width=0.65\linewidth]{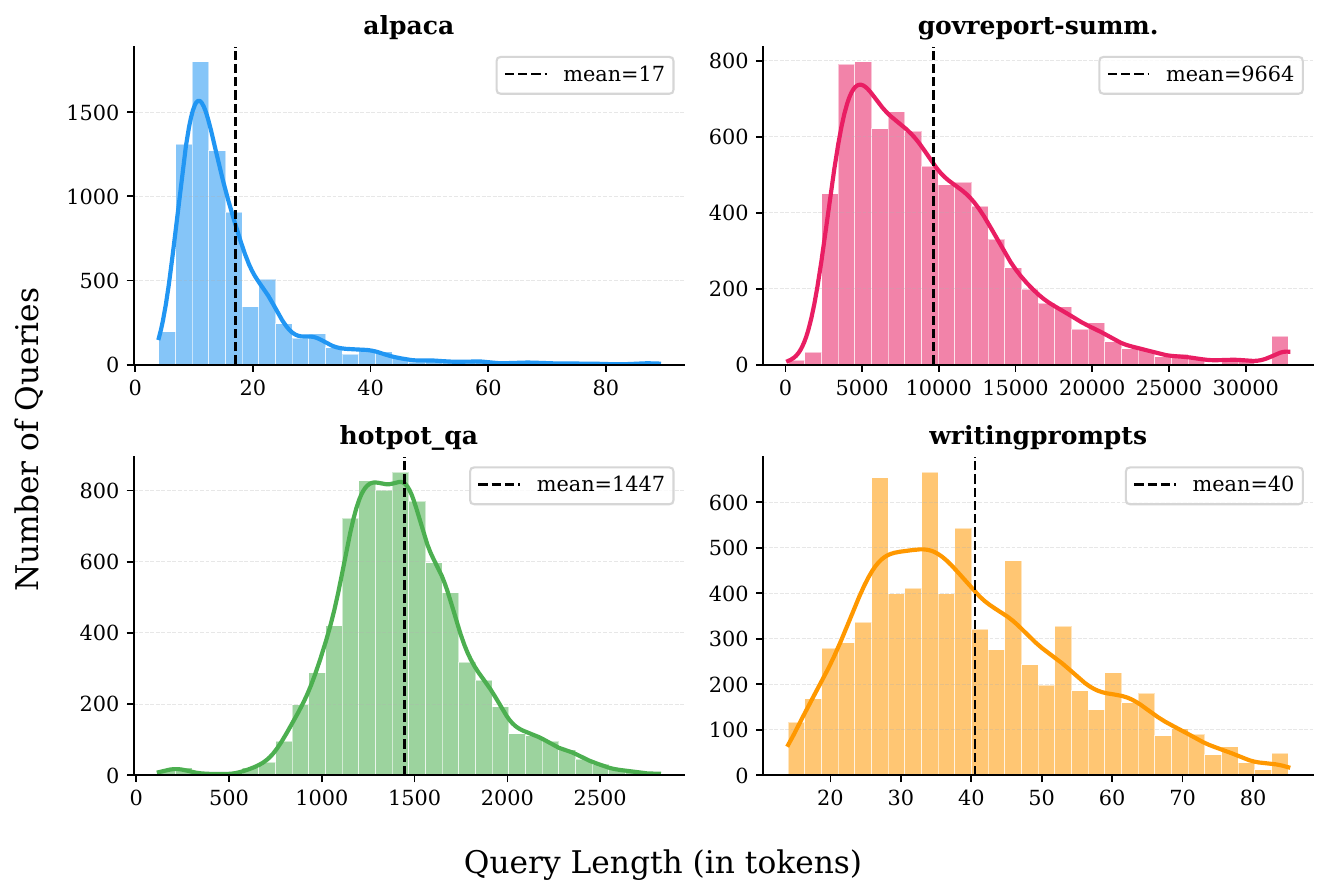}
    \caption{Query token length distributions across the tasks in our evaluations. Alpaca \cite{alpaca_dataset}, GovReport-Summarization \cite{huang-etal-2021-efficient_govreport_summarization}, HotpotQA \cite{yang2018hotpotqa}, and WritingPrompts \cite{fan-etal-2018-hierarchical_writing_prompts_dataset} tasks present a diverse range of input lengths ($\sim10\text{ to }10^4$ tokens), simulating real workloads and a comprehensive evaluation of our routing scheme and latency estimator.}
    \label{fig:query_token_length_distribution}
\end{figure}

\begin{figure}[h]
    \centering
    \includegraphics[width=0.8\linewidth]{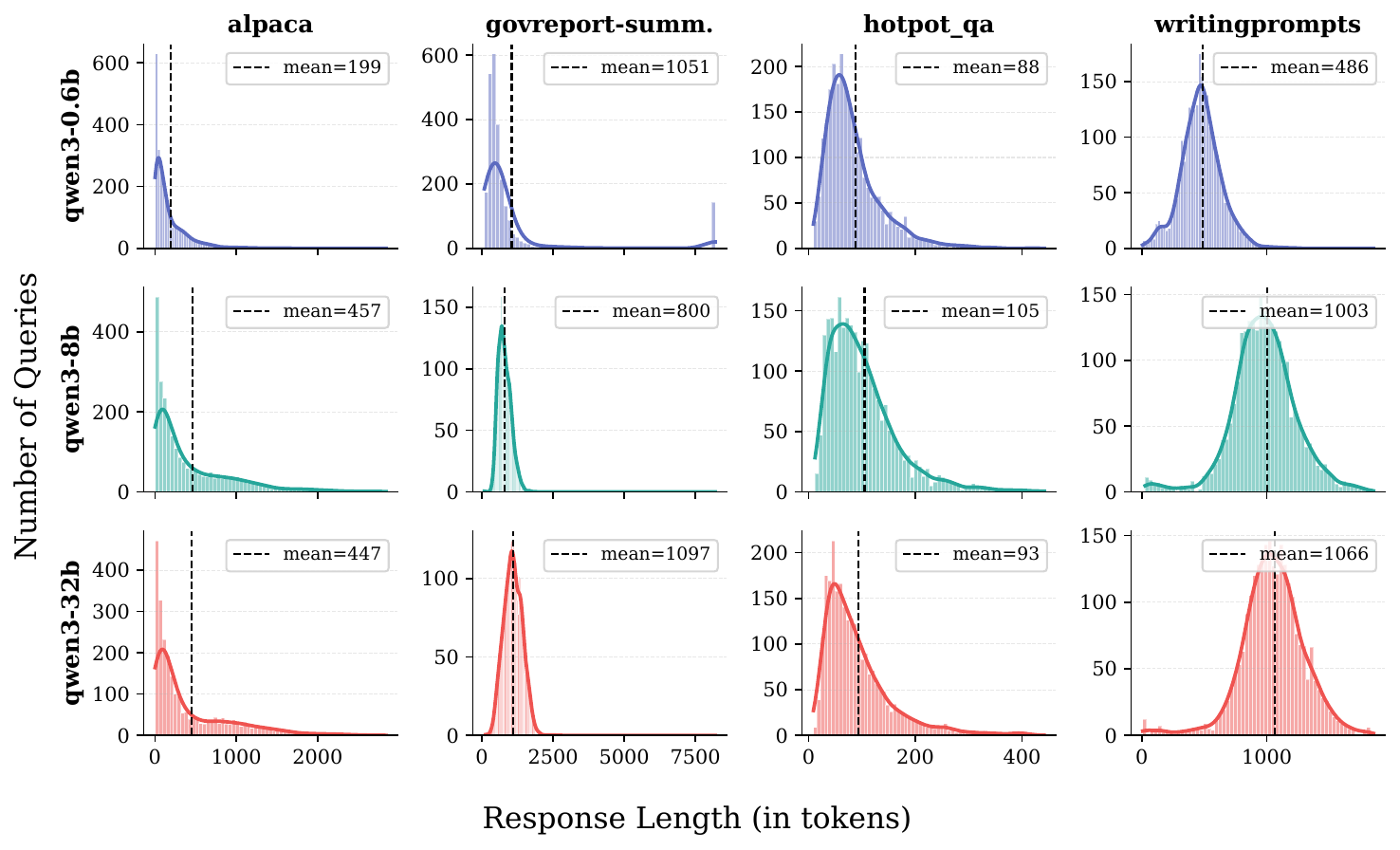}
    \caption{Response token lengths for examined Qwen3-0.6B/8B/32B language models \cite{yang2025qwen3technicalreport} across tasks. Response lengths vary based on the nature of the task (creative generation, summarization, etc.), covering a broad range of $\sim10^2\text{ to }10^3$ tokens.}
    \label{fig:response_token_length_distribution}
\end{figure}

We present the query token length distributions across the tasks in \Cref{fig:query_token_length_distribution}, and the model response length distributions in \Cref{fig:response_token_length_distribution}. The examined tasks and their responses present a wide range of workloads for a comprehensive evaluation of the presented latency estimators and routing scheme.

\newpage
\subsection{Accuracy, cost, and output length predictors}
\label{subapp:acc_output_length_predictors}

We utilize lightweight predictors for estimating model accuracy and output length for queries on arrival. We use LightGBM \cite{NIPS2017_6449f44a_lightgbm} regressors over features that can be computed directly from the prompt, without requiring additional sentence embedding models, to capture query semantics in a fixed-dimensional space. We rely on prompt tokens, and map them into a 512-dimensional signed hashing vector, normalize by token count, and project the vector to 16 dimensions using PCA down-projection calibrated on prompts from the initial modeling phase. These PCA components are concatenated with prompt-derived features such as prompt token count, character count, sentence count, average characters per token, length-constraint indicators, and task-type indicators (such as list, summarize, or explain).

The feature vector for the accuracy estimator contains the prompt features, a categorical model-id feature, and model characteristics such as logarithm of number of model parameters and maximum context length. We train a LightGBM \cite{NIPS2017_6449f44a_lightgbm} Huber regressor on quality scores generated by an LLM judge as described in \Cref{app:llm-as-judge}. 
The output-length predictor estimates the expected number of tokens in the generated response $\widehat{\textrm{tok}}^{\textrm{dec}}_{i,j}$ using the same prompt features and categorical model-id feature as the accuracy predictor. We train a LightGBM mean head directly on completion-token counts, and the runtime predictor uses this mean estimate for routing and cost prediction.

Given per-model prompt-token and output-token prices $\text{cost}^{\textrm{in}}_j$ and $\text{cost}^{\textrm{out}}_j$ in \Cref{tab:qwen3_token_costs}, the predicted request cost is derived from the output-length prediction:
\[
\widehat{\text{cost}}_{i,j} =
\textrm{tok}^{\textrm{pre}}_{q_i,j} \cdot \text{cost}^{\textrm{in}}_j +
\widehat{\textrm{tok}}^{\textrm{dec}}_{q_i,j} \cdot \text{cost}^{\textrm{out}}_j,
\]
where $\textrm{tok}^{\textrm{pre}}_{q_i,j}$ denotes the number of prefill tokens based on tokenization scheme at model instance $j$. These estimates are used for calculating utility in \cref{eq:utility}.

\section{Modeling correlated query arrivals}
\label{app:query_arrival_distribution_ablation}

We evaluate robustness to correlated/bursty query arrivals using a two-state Markov-modulated Poisson process (MMPP-2). In an MMPP-2, arrivals are Poisson conditioned on a latent continuous-time Markov state $Z(t) \in \{L,H\}$, where $L$ is a low-rate state and $H$ is a high-rate burst state \cite{fischer1993mmpp}. This model preserves a fixed long-run mean arrival rate while introducing temporally correlated periods of low and high load, making it suitable for bursty, time-varying request traffic.
For each average arrival rate $\bar{\alpha}\in\{6,7,8\}$ qps, we parameterize the MMPP-2 by the rate ratio $r=\alpha_H/\alpha_L$, the high-state fraction $p_H=0.2$, and correlation time $\tau=2$s. Thus,
\[
\alpha_L = \frac{\bar{\alpha}}{(1-p_H)+p_H r}, \qquad
\alpha_H = r\alpha_L.
\]
The Markov transition rates are $z_{L\rightarrow H}=0.1$ and $z_{H\rightarrow L}=0.4$ per second, so the process spends 80\% of time in the low-rate state and 20\% in the high-rate state. We use homogeneous Poisson arrivals as a low-burstiness baseline, MMPP-2 with $r=3$ as moderate burstiness, and MMPP-2 with $r=6$ as high burstiness. \Cref{fig:query_arrival_distribution_ablation} shows the performance of \bear, our proposed latency-aware router utilizing the \bea latency estimator, against examined baselines.

\begin{table}[h]
\caption{MMPP-2 arrival rates used in the arrival distribution ablation.}
\centering
\begin{tabular}{c c c c}
\toprule
Mean QPS $\bar{\alpha}$ & Arrival setting & Low-state QPS $\alpha_L$ & High-state QPS $\alpha_H$ \\
\midrule
6 & $r=3$ & 4.29 & 12.86 \\
7 & $r=3$ & 5.00 & 15.00 \\
8 & $r=3$ & 5.71 & 17.14 \\
6 & $r=6$ & 3.00 & 18.00 \\
7 & $r=6$ & 3.50 & 21.00 \\
8 & $r=6$ & 4.00 & 24.00 \\
\bottomrule
\end{tabular}
\label{tab:mmpp2-arrival-rates}
\end{table}

\newpage
\section{Response evaluation using LLM-as-a-judge}
\label{app:llm-as-judge}
We utilize the Gemini 3.1 Pro Preview \cite{geminiteam2025geminifamilyhighlycapable} LLM via the Gemini API to evaluate the responses generated by the models in \Cref{sec:experiments}. The evaluation prompt is provided below. We obtain raw scores in the 0-10 range and normalize them to $[0,1]$. 

\begin{promptbox}
\noindent You are a strict evaluator.\par

\medskip
\noindent Evaluate each candidate response against the REFERENCE RESPONSE for the given PROMPT.\par

\medskip
\noindent Multiple candidate responses are provided for the same prompt. Score each one independently, but use the other candidates as comparison context to better calibrate quality differences.\par

\medskip
\noindent Focus on correctness and completeness. Ignore style, tone, and phrasing unless they affect meaning, accuracy, or instruction-following.\par

\medskip
\noindent Penalize contradictions, hallucinations, unsupported claims, and missing key points.\par

\medskip
\noindent Use the full 0 to 10 scale and assign scores granularly across the range to capture meaningful differences in quality.\par

\medskip
\noindent Do not bunch most responses into a narrow band.\par

\medskip
\noindent Use decimals when useful.\par

\bigskip
\noindent Scoring anchors:\par

\medskip
\noindent\hspace*{2em}0 = completely wrong, irrelevant, or unusable\par
\noindent\hspace*{2em}2 = mostly wrong with very little useful content\par
\noindent\hspace*{2em}4 = partially correct but with major mistakes or omissions\par
\noindent\hspace*{2em}6 = substantially correct but missing important details or containing minor errors\par
\noindent\hspace*{2em}8 = strong, correct, and mostly complete with only small gaps\par
\noindent\hspace*{2em}10 = fully correct, complete, and faithful to the prompt and reference\par

\bigskip
\noindent Choose any value from 0 to 10, including decimal values, based on where each response falls between these anchors.\par

\medskip
\noindent Scores should reflect absolute quality while preserving real differences between candidates answering the same prompt.\par

\medskip
\noindent If two responses are extremely similar in quality, their scores may be close or equal.\par

\medskip
\noindent If one response is clearly better or worse, reflect that in the scores.\par

\bigskip
\noindent Return ONLY valid JSON.\par

\medskip
\noindent The JSON must contain exactly one top-level key, \texttt{"scores"}.\par

\medskip
\noindent The \texttt{"scores"} object must contain exactly these keys: \texttt{"A"}, \texttt{"B"}, and \texttt{"C"}.\par

\medskip
\noindent Use this exact structure:\par

\medskip
\noindent\texttt{\{}\par
\noindent\hspace*{2em}\texttt{"scores": \{}\par
\noindent\hspace*{4em}\texttt{"A": number from 0 to 10,}\par
\noindent\hspace*{4em}\texttt{"B": number from 0 to 10,}\par
\noindent\hspace*{4em}\texttt{"C": number from 0 to 10}\par
\noindent\hspace*{2em}\texttt{\}}\par
\noindent\texttt{\}}\par

\bigskip
\noindent DO NOT return any other text since this is for an evaluation task.\par
\end{promptbox}

\section{Beyond Poisson arrivals and exponential/geometric service}
\label{app:lps_general_approx}
For general response generation time distributions, the time until the next admission into generation depends on the \emph{residual} decode sequence lengths of the $k_j$ active jobs, requiring additional analysis beyond the standard Poisson arrivals and exponentially distributed service requirements.
The LPS waiting time expression in \cref{eq:lps_mean_wait} admits a variability-corrected heuristic approximation \cite{whitt_approximations,Kingman_1961_approximations} which we utilize to generalize our examination to general arrival processes and query distributions. Let $Q$ denote the per-query service requirement, with mean $\mathbb{E}[Q]$, and let $\rho_j = \alpha_j \mathbb{E}[Q]$, and squared coefficients of variation of inter-arrival times and service requirement as $c_a^2$ and $c_s^2$ respectively. The corresponding approximation is 
\begin{equation}
\label{eq:lps_general_approx}
\mathbb{E}[W_{i,j}]
\approx
\frac{c_a^2+c_s^2}{2}\cdot
\frac{\rho_j^{k_j}\,\mathbb{E}[Q]}{1-\rho_j}.
\end{equation}
\Cref{eq:lps_general_approx} can be viewed as a first-order extension of the exact LPS formula that accounts for arrival and service variability. In settings where the output-length distribution is strongly heavy-tailed or the arrival process is bursty, a more accurate alternative is to estimate queueing delay by direct simulation or through more complex analytical representations of per query service requirement.

\section{Implementation of the BEA latency estimator}
\label{app:batching_simulation_implementation}

We describe the implementation of the
\bea latency estimator in \Cref{subsec:sfs-latency-estimator}.
Estimation is initiated by the router and uses periodically published query workload snapshots from model instances to predict the time-to-first-token (TTFT) latency a new request would experience at each candidate model instance. A desirable estimator should be computationally efficient and should not induce significant overhead for query routing.

\parahead{Query workload snapshots.}
Each model instance captures and conveys to the router a compact summary of its current query computation workload after each token iteration through shared memory for computational efficiency. This snapshot contains the information utilized by the latency simulator at the router: the resident requests at the instance, their remaining prefill computation workload, predicted decode work, and the scheduler state needed for constructing token batches.

\parahead{Robustness to incomplete snapshots.}
Since query workload snapshots are updated without pausing autoregressive generation at model instances, the router must avoid using partially written snapshots for simulation. We use a simple consistency mechanism in which each snapshot update is marked as either in-progress or complete. The router accepts a snapshot only when it observes that the update has completed and that the snapshot was not modified during the read. This allows the router to obtain a valid view of the instance state without blocking the serving engine.
The router maintains a local cached copy of the most recent complete snapshot from each model instance, which is utilized for latency estimation upon arrival of new queries. 

\parahead{Latency estimation.}
When a new request \(q_i\) arrives, the router evaluates each candidate model instance \(j\) by simulating the batching and scheduling policies of the serving engine given the current workload, along with the new query. The resulting sequence of batches until the first output token generation of the new query (\cref{eq:first_output_token_batch}) determines the TTFT for $q_i$. We utilize a calibrated estimator (\cref{eq:batch_processing_time_estimator}) to compute batch processing times, yielding the TTFT estimator (\cref{eq:sfs_ttft_sum}). This simulation accounts for queue order, admission limits, token-budget constraints, context limits, KV-cache allocation state, and preemption behavior when applicable.

\parahead{Code structure.}
The implementation separates three components: the serving engine, workload snapshot collection, and routing-time evaluation. Each model instance periodically records its current query workload, and the router keeps a cached copy of the latest snapshot from each instance, using them for TTFT latency simulation when a new request arrives. The main routing logic, request dispatch, and policy evaluation are implemented using Python. Critical components such as snapshot parsing and batching policy simulation are implemented in C++ and exposed to Python endpoints. The simulation overhead (involving a loop over request states for construction of each token batch) for latency estimation is minimized through our C++ implementation.

The snapshot processing and scheduling simulation module consists of \(\approx\!3\)K lines of C++, including \(\approx\!300\) lines for the token batch construction loop for the serving engine. Adapting the estimator to another serving engine or scheduling policy only requires modifying this batching simulation logic with minor modifications to router integration.

\section{Additional results}
\label{app:additional_results}

\begin{figure}[h]
    \centering
    \includegraphics[width=0.5\linewidth]{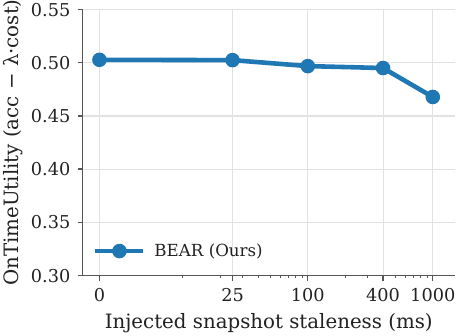}
    \caption{\small \textbf{Robustness to workload snapshot staleness.}
    OnTimeUtility of \bear under snapshot delays, simulating delayed communication between model instances and the router. OnTimeUtility performance remains close to the zero-staleness value for delays up to $400$\,ms, with only a $7\%$ decrease at $1000$\,ms.}
    \label{fig:snapshot_staleness}
\end{figure}

\parahead{Robustness to workload snapshot staleness.}
In distributed deployments, communication delays can cause workload snapshots to become stale before they are used for latency estimation and routing. Queries at model instances may arrive, progress, or complete during this interval, potentially affecting TTFT estimates and routing decision.
To evaluate this sensitivity, we inject snapshot staleness of $0$, $25$, $100$, $400$, and $1000$\,ms and measure the resulting OnTimeUtility (at Poisson arrival with $8$ queries per second). We build latency estimates using the delayed workload information, while the model instances continue serving existing and potentially new requests.
\Cref{fig:snapshot_staleness} shows that OnTimeUtility remains close to the zero-staleness value for delays up to $400$\,ms, decreasing by only $7\%$ even at $1000$\,ms staleness.
Thus, moderate snapshot delays have a limited effect on the routing performance and objective values under the evaluated workload.
Although staleness in workload information can change predicted latencies, such changes need not alter the feasible model instance set or the selected instance. 
The observed stability is consistent with this behavior, indicating that our latency-aware routing policy can tolerate moderate communication delays without requiring synchronization with every token-generation iteration. 

\begin{figure}[h]
    \centering
    \includegraphics[width=0.99\linewidth]{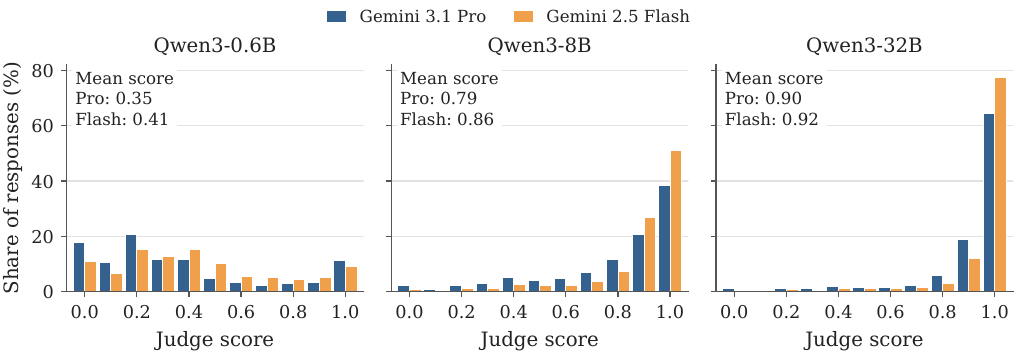}
    \caption{\small \textbf{Sensitivity of response-quality scores to judge choice.}
    Gemini 3.1 Pro and Gemini 2.5 Flash preserve the aggregate model ordering, with model-wise mean scores annotated in each panel. Across query evaluations, the judges agree on pairwise model ordering in $83.19\%$ of cases and share the top-scoring model for $86.64\%$ of queries.}
    \label{fig:judge_score_distributions}
\end{figure}

\parahead{Sensitivity to LLM-as-a-judge choice.}
To assess the sensitivity of the evaluated response quality to the choice of LLM-as-a-judge \cite{li-etal-2025-generation_llm-as-judge}, we compare evaluations from Gemini 3.1 Pro and Gemini 2.5 Flash (scores normalized to $[0,1]$ range).
\Cref{fig:judge_score_distributions} shows that both judges broadly preserve the aggregate ordering of the language models under consideration. 
The considered judges agree on the relative ordering of model pairs for $83.19\%$ of query--model-pair evaluations, and they also share at least one top-scoring model for $86.64\%$ of queries.
These results suggest that the choice of LLM judge has a limited impact on the relative response quality signal used to train our accuracy estimator.

\begin{figure}[h]
    \centering
    \includegraphics[width=0.55\linewidth]{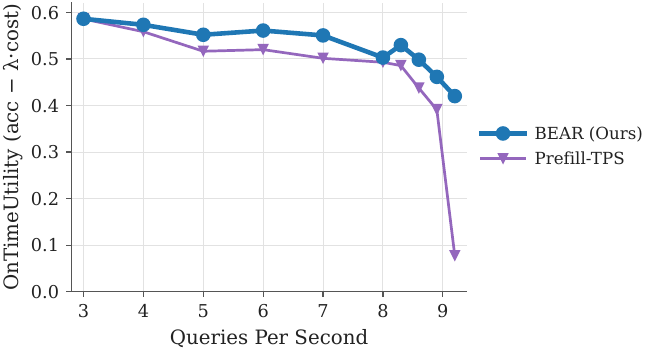}
    \caption{\small \textbf{Latency-estimator ablation for latency-constrained routing.}
    Both variants use identical response-quality and cost predictors and the same routing rule, differing only in the TTFT estimator. The \bea estimator achieves higher OnTimeUtility than the prefill throughput-based latency estimator as load increases, with a higher advantage at increasing query arrival rates.}
    \label{fig:prefill_throughput_ablation}
\end{figure}

\parahead{Comparing latency estimators.}
We complement the TTFT prediction comparison in \Cref{fig:ttft_estimator_comparison} with an ablation examining how latency estimator accuracy affects downstream routing performance.
We retain the same response-quality and cost predictors, and the latency-constrained routing objective in \cref{eq:constrained_routing_objective}, replacing only our latency estimator with the prefill throughput-based latency estimator (\emph{Prefill-TPS}).
Prefill-TPS utilizes the remaining prefill backlog and incoming prompt length to estimate latency using the model instance's measured prefill throughput.
Our latency estimator instead simulates inference and token-batch evolution under the serving framework's scheduling policies, allowing it to account for prefill--decode interference, KV-cache memory constraints, and memory-dependent request admission. 
\Cref{fig:prefill_throughput_ablation} shows comparable OnTimeUtility at the low arrival rates, but higher performance with our latency estimator as the load increases.
Together, \Cref{fig:ttft_estimator_comparison,fig:prefill_throughput_ablation} show that improved TTFT prediction translates into better routing performance under congestion, with the accuracy--cost predictors and routing objective held fixed.

\begin{figure}[h]
    \centering
    \begin{minipage}[t]{0.47\textwidth}
        \centering
        \includegraphics[width=\linewidth]{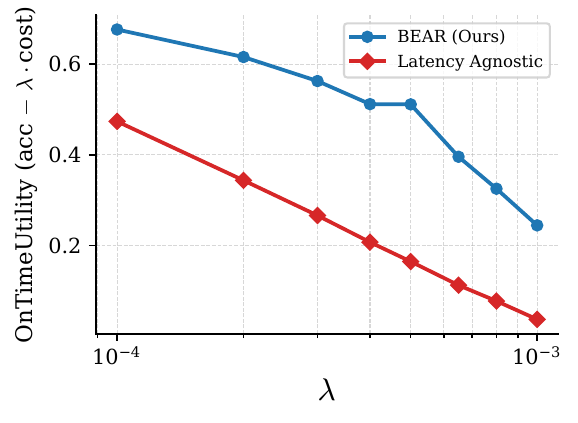}
        \caption{\small OnTimeUtility with varying accuracy--cost balance ($\lambda$) for our approach and latency agnostic routing. Consistent improvement across varying accuracy--cost balances indicates the generalizability of our approach.}
        \label{fig:lambda_sweep_results}
    \end{minipage}
    \hfill
    \begin{minipage}[t]{0.51\textwidth}
        \centering
        \includegraphics[width=\linewidth]{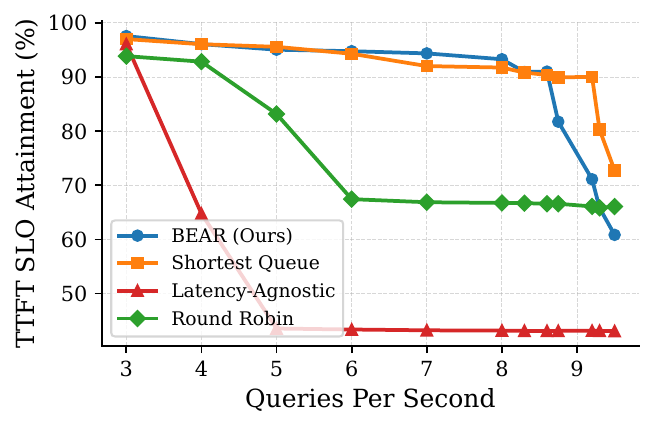}
        \caption{\small Latency constraint (SLO) attainment vs. query arrival rate. Our approach makes reliable latency estimates for queries across model instances, resulting in high latency constraint satisfaction values.}
        \label{fig:slo_attainment_vs_qps}
    \end{minipage}
\end{figure}

\parahead{Varying $\lambda$ in utility \cref{eq:utility}.}\Cref{fig:lambda_sweep_results} denotes the OnTimeUtility of our approach as compared to \emph{Latency-Agnostic} policy (\emph{Shortest Queue} and \emph{Round Robin} do not have $\lambda$ parameter to control the accuracy--cost tradeoff). We observe that across different choices of $\lambda$ values, our approach maintains higher performance. Note that the utility values naturally reduce as $\lambda$ increases in \cref{eq:utility}.

\parahead{Query latency constraint attainment.} We present the query latency attainment in \Cref{fig:slo_attainment_vs_qps}, which denotes that our approach maintains high SLO attainment for queries through accurate estimates of latencies across model instances that match with observed latencies. Combined with improved accuracy--cost utility, our approach shows the highest OnTimeUtility across all baselines as demonstrated in \Cref{fig:ontimeutility_vs_qps}.

\begin{figure}
    \centering
    \includegraphics[width=0.95\linewidth]{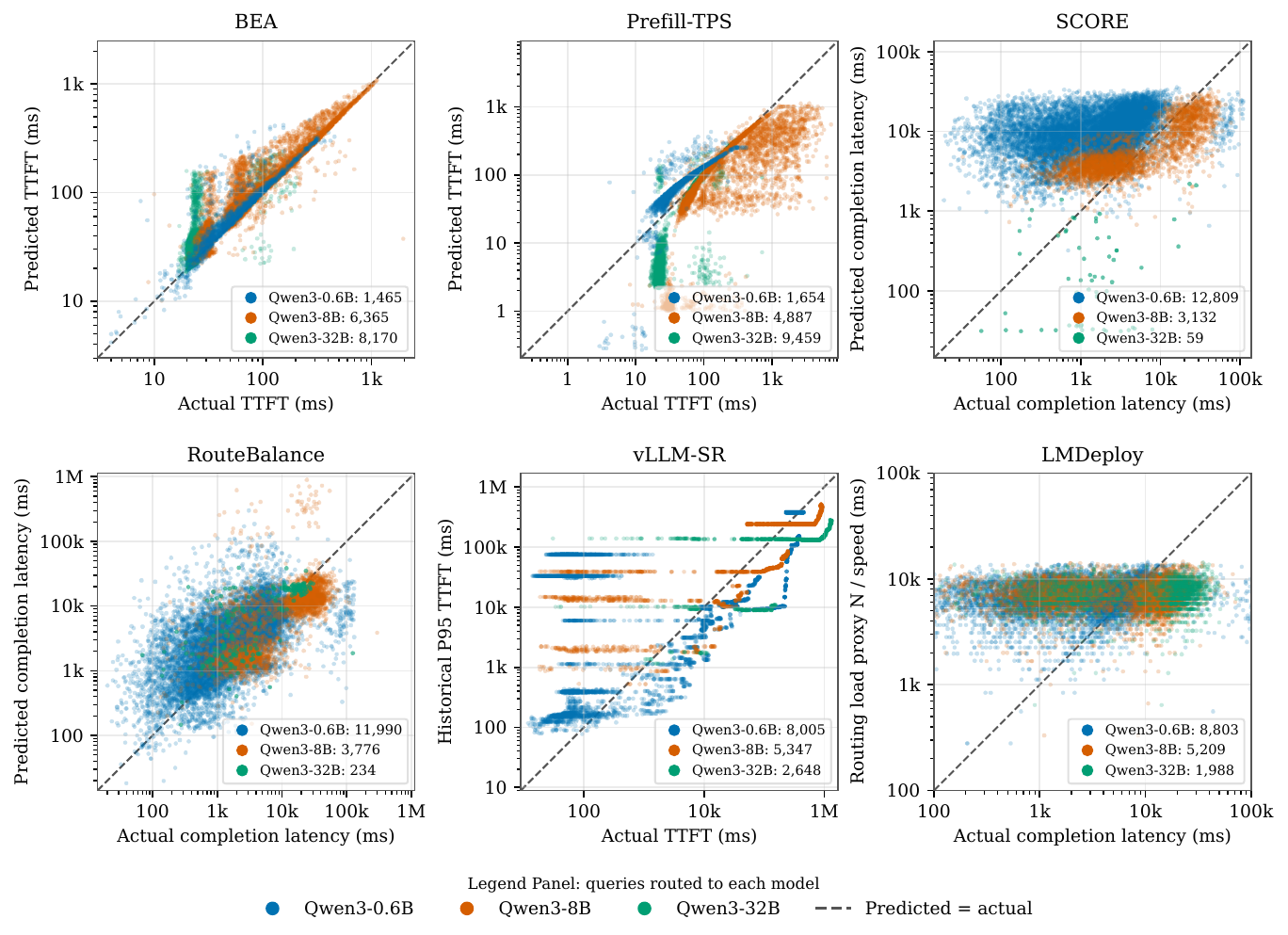}
    \caption{Latency estimates for \bea and examined baselines across Qwen3-0.6B, Qwen3-8B, and Qwen3-32B models. \bea uses batch processing-time estimates that account for varying prefill--decode compositions, enabling accurate latency predictions as serving workloads evolve.
}
    \label{fig:latency_estimation_comparison}
\end{figure}

\parahead{Latency Estimation across Baselines} We examine the latency estimation performance of \bea and the baselines, to help contextualize the gains observed in routing performance (\cref{fig:ontimeutility_vs_qps,fig:utility_vs_latency}). SCORE \cite{lakha2025faster_cheaper_justasgood} utilizes a measured time per output token and estimated remaining token workload to predict latencies, whereas RouteBalance \cite{da2026routebalancefusedmodelrouting} employs the ratio of number of pending decode tokens across all queries and active decode batch size to estimate time to completion along with time per output token predictions. Both these approaches estimate latencies based on estimates of decode lengths for each sequence, leading to decode length estimation errors being translated to latency estimation errors. Prefill throughput based estimator (\Cref{fig:ttft_estimator_comparison}) relies on a pre-calculated prefill throughput estimate which fails to account for varying prefill--decode batch compositions during inference. Furthermore, LMDeploy's \cite{zhang2025lmdeploy} latency estimation calculates the ratio of unfinished requests and request throughput to obtain completion latency predictions, resulting in poorly calibrated predictions across varying query lengths. vLLM-SR \cite{liu2026vllmsemanticroutersignal} considers normalized percentile values of previously observed latencies to estimate new query's TTFT at model instances. The percentile values are calculated over a rolling window of past observations, leading to poor real-time latency estimates as workloads at model instances evolve with time. 

\end{document}